\def\year{2020}\relax
\documentclass[letterpaper]{article} 
\usepackage{aaai20}  
\usepackage{times}  
\usepackage{helvet} 
\usepackage{courier}  
\usepackage[hyphens]{url}  
\usepackage{graphicx} 
\usepackage{algorithm,algorithmic}
\usepackage{graphicx}  
\newcommand{\citet}[1]{\citeauthor{#1} \shortcite{#1}}
\newcommand{\citep}{\cite}

\usepackage{amsmath,amssymb}

\usepackage{multirow}
\usepackage{subcaption}
\usepackage{booktabs}

\nocopyright
\title{Detecting Out-of-Distribution Inputs in Deep Neural Networks\\ Using an Early-Layer Output}
\author{ 
}

\author{
 Vahdat Abdelzad\thanks {Department of Electrical and Computer Engineering, University of Waterloo}\\
 vabdelza@uwaterloo.ca
 \And
 Krzysztof Czarnecki\textsuperscript{\rm *}\\
 kczarnec@gsd.uwaterloo.ca \\
 \And
 Rick Salay\textsuperscript{\rm *}\\
  rsalay@gsd.uwaterloo.ca\\
 \AND
  Taylor Denounden\thanks{Department of Computer Science, University of Waterloo}\\
  tadenoud@uwaterloo.ca\\
 \And
  Sachin Vernekar\textsuperscript{\rm $\dagger$}\\
  sverneka@uwaterloo.ca\\
  \And
  Buu Phan\textsuperscript{\rm *}\\
  btphan@uwaterloo.ca\\
}

\begin{document}

\maketitle

\begin{abstract}
Deep neural networks achieve superior performance in challenging tasks such as image classification. However, deep classifiers tend to incorrectly classify out-of-distribution (OOD) inputs, which are inputs that do not belong to the classifier training distribution. Several approaches have been proposed to detect OOD inputs, but the detection task is still an ongoing challenge. In this paper, we propose a new OOD detection approach that can be easily applied to an existing classifier and does not need to have access to OOD samples. The detector is a one-class classifier trained on the output of an early layer of the original classifier fed with its original training set. We apply our approach to several low- and high-dimensional datasets and compare it to the state-of-the-art detection approaches. Our approach achieves substantially better results over multiple metrics.
\end{abstract}

\begin{figure*}
\centering
\subcaptionbox{\label{id_ood_umap_1:ma}}{\includegraphics[width=4.3cm]{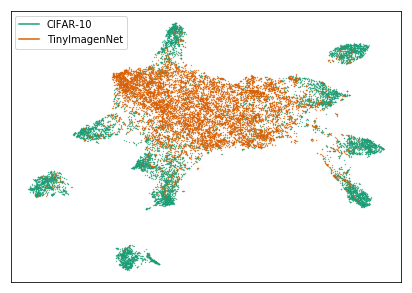}}
\subcaptionbox{\label{id_ood_umap_1:mb}}{\includegraphics[width=4.3cm]{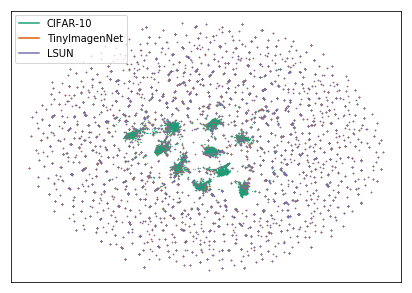}}
\subcaptionbox{\label{id_ood_umap_1:mc}}{\includegraphics[width=4.3cm]{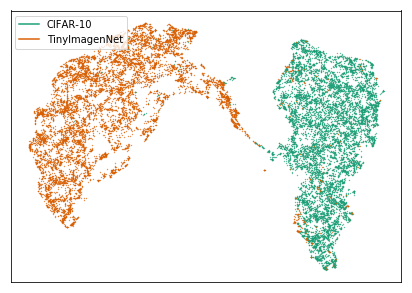}}
\subcaptionbox{\label{id_ood_umap_1:md}}{\includegraphics[width=4.3cm]{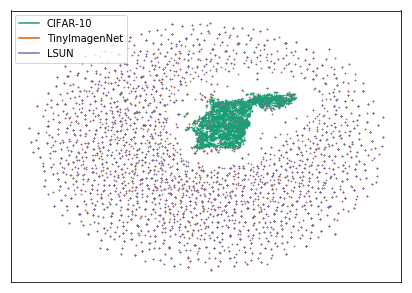}}

\caption{ Two-dimensional representations of features extracted from a ResNet model trained on CIFAR-10. a) Features extracted from the penultimate layer for the test set of CIFAR-10 and TinyImageNet datasets. b) Similar to (a) but LSUN was added as the second OOD dataset. c) Similar to (a) but features have been extracted from a well-chosen layer. d) Similar to (b) but features have been extracted from a well-chosen layer. Visualizations use UMAP \citep{umap_2018}. }  
\label{id_ood_umap_1}
\end{figure*}

\section{Introduction} \label{intoduction}
\noindent Deep Neural Networks (DNNs) are an indispensable part of the current generation of software systems for Autonomous Vehicles (AV), the Internet of Things (IoT), and medical diagnosis. We can observe the power of DNNs specifically in the advanced deep models developed for vision and speech recognition systems~\cite{waveNetGoogle2016,alexConv2012,resnet_He_2015}. Classification is one of the tasks that humans perform regularly and deep models sometimes outperform humans in doing this task. 

Deep classifiers generalize well when they are given inputs drawn from the same distribution as the training data, referred to as in-distribution (ID). In practice, inputs can be drawn from the in-distribution or other distributions, however. Modern deep networks tend to predict such out-of-distribution (OOD) inputs as an ID class with high confidence \cite{deep_fool_2014}.

This misbehavior can hinder the adoption of deep classifiers in safety-critical systems. For example, a classifier trained to classify vehicles may misclassify a vulnerable road user on a recumbent bike, if not in the training set, as a car. Therefore, classifiers need to be enhanced with mechanisms that allow distinguishing ID and OOD inputs. The problem of detecting OOD inputs has been studied extensively in different domains under various names such as outlier and novelty detection \cite{Pimentel-novelty-2014,Chandola-Anomaly-2009}.

There are several approaches proposed to detect OOD inputs for deep models~\citep{Hendrycks-baseline-2017,devries2018learning,Liang-ODIN-2018,Lee-mah-2019}. The baseline approach is max-softmax proposed by \citet{Hendrycks-baseline-2017}, which uses a threshold over the predicted softmax class probability to detect OOD inputs. This approach does not put any assumption over the architecture of deep models or use samples of OOD inputs for detection. It can also be applied to already trained models. However, it does not have satisfactory performance. Other approaches may have better performance than max-softmax, but they constrain the architecture of deep models, put assumptions over the distribution of deep features, cannot be applied to already trained models, or need samples of OOD inputs in order to have better performance \citep{devries2018learning,Liang-ODIN-2018,Lee-mah-2019}. Therefore, providing a detection approach which is free of those constraints and outperforms max-softmax with a good margin is still a challenge. In this paper, we concentrate specifically on detection approaches that do not require the classifier to be retrained or re-designed, since deep classifiers may be very costly to train, or have specific constraints over architecture.

Although the aforementioned approaches use different techniques to detect OOD inputs, they all essentially rely on features extracted by the penultimate layer of deep classifiers. This layer has been trained to extract features that are important to separate the ID classes of inputs. Figure \ref{id_ood_umap_1:ma} shows a two-dimensional representation of the features extracted from the penultimate layer of a ResNet model trained on CIFAR-10 \citep{cifar_2009}. The features belong to the test datasets of CIFAR-10 and TinyImageNet \citep{imagenet_2009}, which are considered ID and OOD, respectively. There are ten small clusters (i.e, manifolds) of features that represent different classes (i.e., ID inputs) of CIFAR-10. There is also a large cluster of features that represents OOD inputs. In this case, the detection problem manifests itself as separating the distribution of OOD features from the distributions of the different classes (i.e., ten different distributions). This can be challenging even when a distance function or a learning method is used to separate them. Figure \ref{id_ood_umap_1:mb} depicts the features extracted for CIFAR-10, TinyImageNet, and LSUN datasets \citep{LSUN_2015}. Here, it becomes much harder to separate ID and OOD inputs.
\citet{Hein_2018_Relu} also demonstrate that softmax-based approaches cannot avoid assigning a high-class probability for inputs that are far from the training distribution (i.e., OOD inputs). It is deemed as an inherent problem of DNNs.


We propose a new approach that is not based on class probability (the output of the softmax layer), it significantly outperforms the-state-of-the-art approaches, does not need OOD samples for training the detection model, and does not require retraining the classifier. Our main insight is that there is a latent space in which the distributions of ID and OOD datasets are well separated (regardless of the distribution of classes) and the transformation function to such a latent space has already been approximated well by one of the classifier's layers. This insight allows learning a detector based on features in this space to separate ID and OOD inputs.

Figure \ref{id_ood_umap_1:mc} shows a two-dimensional representation of features in such a latent space for the same model and datasets used in Figure \ref{id_ood_umap_1:ma}. There are two different clusters, each one representing features of ID and OOD datasets, respectively. Being able to represent the entire ID distribution as a manifold in a space can result in better separation. Figure \ref{id_ood_umap_1:md} shows features in such a space for an ID and two OOD datasets (similar to Figure \ref{id_ood_umap_1:mb}). Here, the ID features are again bounded well whereas OOD features are distributed around. In our approach, we find such a latent space and learn a secondary model (which is trained in a few minutes) to separate distributions of features for ID and OOD datasets. We apply our approach to several low- and high-dimensional datasets and then compare it to the baseline and state-of-the-art approaches. The results demonstrate significant improvement for multiple detection metrics. 

The rest of paper is organized as follow. In Section 2, we briefly go over related work. Then, we introduce our approach in Section 3, including how to detect OOD inputs and how to find suitable latent space. In Section 4, we present our experimental results. Finally, we conclude and suggest directions for future work in Section 5.

\section{Related work}

\citet{Hendrycks-baseline-2017} propose a baseline approach to detect OOD inputs, called max-softmax, and a set of metrics to evaluate OOD detectors. The approach relies on the observation that ID inputs tend to have a higher predicted softmax class probability compared to OOD inputs. Therefore, a threshold over the predicted softmax class probability should allow separating ID and OOD inputs. Despite being a simple and easy-to-use approach to detect OOD inputs, the approach does not have satisfactory performance, especially for critical applications.

\citet{Liang-ODIN-2018} propose an approach called ODIN that improves on max-softmax by incorporating two extra components: temperature scaling and input preprocessing. Temperature scaling \citep{Guo-Temperatue-2017} is used to calibrate the softmax output of a network, and input preprocessing helps to increase its maximum output for ID inputs. Although preprocessing improves the detection of ID inputs, it needs \textit{access to OOD samples in advance} to fine-tune the perturbation magnitude used in preprocessing during inference. In practice, it is possible to have access to some OOD samples, but it is hard or impossible to have access to samples from all OOD datasets. Input preprocessing also requires extra processing time (two forward and one backward passes over the model), which can be an issue for real-time systems.

The performance of the softmax-based approaches (such as ODIN) depends highly on how the predicted softmax class probability varies for ID and OOD inputs. A higher class probability for an input indicates a higher chance of considering it as ID. \citet{Kimin_Lee-2018} further improve upon ODIN and propose an approach that forces deep classifiers to output close to uniform distribution for OOD samples. The main idea is to jointly train a classifier and a generator: the generator is trained to produce samples at the boundary of the data manifold, which are considered OOD samples; the classifier is trained using a specially designed loss function that encourages the classifier to assign uniform class probabilities to these generated OOD samples. This approach requires expensive retraining for an existing classifier and relies on the assumption that generated samples cover the entire boundary of the data manifold, which is difficult to achieve for high-dimensional data \citep{Confident-Classifiers_2019}.

\citet{devries2018learning} utilize uncertainty to detect OOD inputs. They assume that a classifier is more confident about its prediction when the input is ID. Therefore, they proposed to retrain a classifier to output also confidence estimates for each input. Then, the confidence score is used to differentiate between ID and OOD inputs. Others \cite{Shalev2018,Vyas2018} suggested to use ensembles to calculate confidence estimates. Similar to the approach proposed by \citet{Kimin_Lee-2018}, these approaches cannot be applied to already trained models.

MC-dropout \citep{Gal-dropout-2016} is another technique to measure uncertainties of models. This technique relies on multiple inferences to calculate uncertainties and may not be practical in real-time systems. Futhermore, it can only be applied to models that have been trained with dropout layers. \citet{Geifman-selective-2017} used MC-dropout to draw the risk-coverage curve, which is similar to the concept of OOD detection, for already trained models. They reported that MC-dropout performance is similar to max-softmax. Moreover, uncertainty-based approaches are well-suited for detecting confusing inputs existing near to the boundaries of classes that a model trained for. This affects negatively the effectiveness of these approaches for inputs far from the training
distribution. 

Using generative models is also another way to detect OOD inputs \citep{Denouden_ae_reconstruction_2018,Pidhorskyi_adv_2018}. These approaches usually rely on either the reconstruction error or an estimation of density using the latent representations or a combination of the two. However, these approaches are found to give a higher likelihood to some of the OOD datasets \citep{Nalisnick_2018}. Furthermore, generative models tend to scale poorly with the dimension of the dataset. For example, a reconstruction-based OOD detection approach performs well for MNIST, but it becomes difficult to train a generative model to reliably capture the latent manifold of high dimensional datasets~\citep{Wang_safe_classification2017}.

\citet{Lee-mah-2019} proposed an approach in which they obtain the class conditional Gaussian distributions with respect to features of the deep models under Gaussian discriminant analysis. This results in a confidence score based on Mahalanobis distance (MD) that is used to detect OOD inputs. They propose two approaches (without considering input preprocessing):  a) MD over features before the logits layer; b) an ensemble model based on MDs for all layers. The former performance is similar to ODIN as indicated in their original paper (Table 1 in \citet{Lee-mah-2019}). The latter trains a regression model with OOD samples for each OOD dataset. In fact, for $n$ OOD datasets, $n$ regression models are trained. This is a significant limitation of their approach because the regression models are biased toward the particular OOD distributions on which they are trained. \citet{Lee-mah-2019} also used adversarial samples to train the regression models. This resulted in a reduction in the detection performance (shown by the variation in the performance on the right-hand side of Table 2 in \citet{Lee-mah-2019}), which is a sign of biases. Last, for both approaches (a \& b) proposed by \citet{Lee-mah-2019} distributions of features for each class must follow the multivariate Gaussian distribution. There is no guarantee that distributions of features will satisfy this assumption.

Our approach detailed in Section 3, although seemingly similar to the approach by \citet{Lee-mah-2019} because of using deep features, differs in several important ways: i) it does not use features in the logits layer. We demonstrated in Figures \ref{id_ood_umap_1:mc} and \ref{id_ood_umap_1:md} that features in such a layer are not appropriate to separate ID and OOD; ii) it does not need to have access to OOD samples. \textit{We always train one model (without OOD or adversarial samples) and this single model applies to all OOD datasets}; iii) it finds the appropriate feature space in which ID and OOD inputs are well-separated and thus has no need to rely on ensembles; iv) it does not force the feature distributions to follow the multivariate Gaussian distribution.

\section{Proposed solution}
In this section, we describe different elements of our approach, including how detection is performed for a specific latent space, how such a latent space is found, and how the approach can be enhanced using input preprocessing. 

\subsection{OOD detection}
Let $Q: \mathbb{R}^n \rightarrow [0,1]^c $ be a function representing a deep network (i.e., classifier), where $x\in \mathbb{R}^n$ is the input and $c$ is the number of classes. Denote as $Q_i$ the output of network $Q$ for class $i$, and as $X=\{x_1, \dots, x_m\}$ the training set. Network $Q$ has $L$ layers and the output of layer $l$ (i.e., activation values) is represented by $q^l$ ($q^0 = x$). Indeed, $q^l$ is the representation of input $x$ in a latent space obtained by non-linear transformations from layers $1$ to $l$. Each layer allows extracting unique features related to input $x$. For example, the final layer of a network extracts features that are important to separate the class of input. 

The hypothesis underlying our approach is that there is also a latent space, represented by the output of a layer called \textit{optimal OOD discernment layer (OODL)}, in which represented features are discriminative enough to allow separating distributions of ID and OOD datasets. In particular, we deem that features used to separate the class of input might not be appropriate to decide whether or not an input is OOD. If we learn the probability distribution function of features obtained by the OODL, then it should be feasible to separate ID and OOD inputs. Our experiments presented later confirm this hypothesis for the studied datasets and networks.


\begin{algorithm}
\caption{Finding the OODL for detecting OOD inputs}
\label{algo1}
\begin{algorithmic}[1]
   \REQUIRE $t\_ds$: training dataset set, $id\_ds$: ID dataset, $ood\_ds$: OOD dataset, $\{\textit{nu}, k\}$: training error and kernel for OSVM.
   \STATE Initialize the detection error vector: $E=0$
   \FOR {each layer $l \in1, \dots, L$}
     \STATE{Extract features of layer $l$ for training dataset:\\$Q^l = \textit{extract}\_q(\textit{t\_ds}, l)$}
     \STATE {Train a one-class SVM classifier:\\$S^l = \textit{oneClassSVM.fit}(Q^l, \textit{nu}, k)$}
     \STATE{Extract features of layer $l$ for ID data:\\$\textit{ID}^l = \textit{}\_q(\textit{id}\_ds, l)$}
     \STATE{Extract features of layer $l$ for OOD data:\\ $\textit{OOD}^l = \textit{extract}\_q(\textit{ood\_ds}, l)$}
     \STATE{Calculate the detection error for layer $l$:\\ $E[l] = \textit{cal\_det\_error}(S^l,\textit{ID}^l, \textit{OOD}^l) $ }
   \ENDFOR
   \RETURN $argmin(E)$
\end{algorithmic}
\end{algorithm}

Now, let $l_{o}$ be the \textit{OODL} and $Q^{l_{o}}$ be the output of layer $l_{o}$ for every $x_i \in X$. Then, we could train a classifier named $S^{l_{o}}$ based on $Q^{l_{o}}$ features to separate ID and OOD inputs. However, to train a two-class classifier $S^{l_{o}}$, access is needed to both ID and OOD features. Although there might be some OOD samples available during the training (e.g., related to a specific OOD distribution), the classifier needs to have samples from all OOD distributions in order to be trained well. Furthermore, having access to specific OOD samples during training will cause classifier $S^{l_{o}}$ to be biased toward detecting those OOD inputs.

Therefore, we formulate this as a one-class classification (OCC) problem. In the OCC context, most of the training samples are ID, and the inputs at inference time are expected to include both ID or OOD samples. Thus, there is no need to have OOD samples during training. OCC has been studied extensively \citep{khan_madden_2014,Perera_OCC_2018} and most existing classification methods can be used for this purpose. In this paper, we use One-class Support Vector Machine (OSVM) \citep{TAX19991191}, which is a commonly used one-class classification algorithm. OSVM outputs a score for a given input $x$, and thresholding over that score allows us to detect OOD inputs. Our detection mechanism is defined as follows.

\begin{equation}\label{eq:ood_detection}
O_{l_o}(x; \delta) =
  \begin{cases}
  0 \quad \delta \geq S^{l_o}(  q^{l_o}(x) )  \\[3.5pt]
  1 \quad \textrm{otherwise} \\ 
  \end{cases}
\end{equation}

$O_{l_o}$ is the detection function based on features in the OODL $l_o$ and $\delta$ is the detection threshold. When the output of classifier $S^{l_o}$ for the features of input $x$ in layer $l_o$ is greater than $\delta$, input $x$  is ID, otherwise, it is OOD. When layer $l_o$ is a fully-connected layer, $q^{l_o}$ is the exact output of layer $l_o$. However, when layer $l_o$ is a convolutional layer and $q^{l_o}$ is the exact output of layer $l_o$, $q^{l_o}$ becomes high-dimensional. This can have negative effect on the performance of OSVM, because OSVM performs better for low-dimensional data. Therefore, we calculate the mean of each channel to reduce the dimension of $q^{l_o}$. Precisely, let $f^{l_o} \in  \mathbb{R}^{w\times h \times d}$ be the feature map of convolutional layer $l_o$, where $w$, $h$, and $d$ are width, height, and depth, respectively. Let $f_{ijk}^{l_o}$ be the $(i, j, k)$-th element of $f^{l_o}$, then, $q^{l_o} = (q_k^{l_o}) \in \mathbb{R}^d$ is given by
\begin{equation}\label{eq:norm_cnn2}
q_k^{l_o} = \frac{1}{w \times h} \sum_{i = 1}^{w} \sum_{j = 1}^{h} \big | f_{ijk}^{l_o} \big |.
\end{equation}

\begin{figure*}
\centering
\subcaptionbox{\label{sfig:ma}}{\includegraphics[width=8cm]{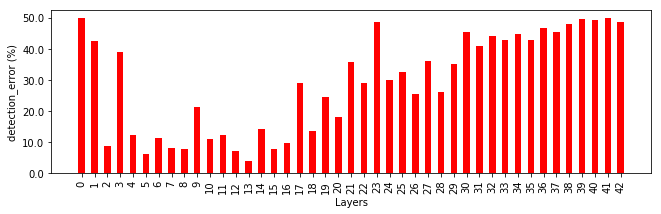}}
\subcaptionbox{\label{sfig:mb}}{\includegraphics[width=8cm]{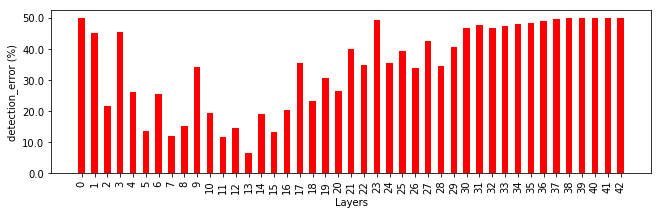}}
\subcaptionbox{\label{sfig:mc}}{\includegraphics[width=8cm]{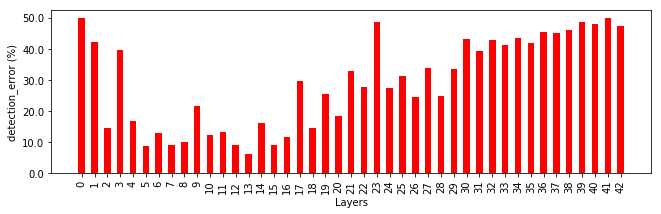}}
\subcaptionbox{\label{sfig:md}}{\includegraphics[width=8cm]{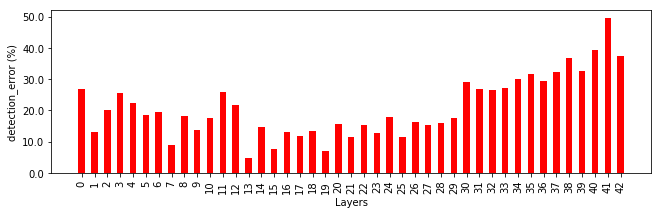}}
\caption{The detection error for residual layers of a ResNet model trained over CIFAR-10. a) The detection error for TinyImageNet. b) The detection error for LSUN. c) The detection error for iSUN. d) The detection error for SVHN. All experiments use the test set of CIFAR-10 as ID dataset.}  
\label{fig:1}
\end{figure*}

Note that traditional OCC methods such as OCSVM perform poorly on high-dimensional data. In our approach, we propose OODL, which allows such methods to perform well on high-dimensional data. An additional benefit is that we can leverage an existing model trained for the original classification task, which is more efficient than training an OCC model from scratch using the ID samples. To the best of our knowledge, no one has proposed using such a setup for OOD detection before.

\subsection{Finding optimal OOD discernment layer (OODL)}
Our approach requires finding the OODL $l_o$ for a given network $Q$. To do so, we use an OOD dataset to measure the detection error (defined in Section \ref{evaluation-metrics}) for layers of network $Q$ based on Equation \eqref{eq:ood_detection}. The layer with the minimum detection error is then selected as the OODL.

Algorithm \ref{algo1} shows how the selection is performed. $t\_ds$ is the training dataset of network $Q$ that is used to extract features for layer $l$. $S^l$ is then trained on the features of layer $l$ according to parameters \textit{nu} and $k$ that are training error and kernel type, respectively. \textit{id\_ds} and \textit{ood\_ds} are ID and OOD datasets used to measure the detection error based on $S^l$. \textit{id\_ds} is chosen to be the test set of network $Q$, and \textit{ood\_ds} can be an arbitrary OOD dataset.

Using an OOD dataset to find the OODL might indicate that such a layer would be biased toward features allowing a better separation of OOD inputs coming from the selected OOD dataset. However, we experimentally found that the choice of an OOD dataset does not affect the approximate position of the OODL. For example, Figure \ref{fig:1} shows the detection error measured for residual layers of a ResNet model trained on CIFAR-10 for different OOD datasets (TinyImageNet, LSUN, iSUN\citep{iSUN_2015}, and SVHN \citep{svhn_2011}). As seen, the OODL  stays the same ($l_o=13$) regardless of which OOD dataset is chosen to measure the detection error. \textit{Also note that the OOD dataset is not used during training $S^{l_o}$}. It is used only to calculate the detection error to select the OODL.

However, the OODL varies based on the ID dataset and deep model. For example, a ResNet model trained on CIFAR-10 has a different OODL than one trained on CIFAR-100 even though the architecture is fixed. Furthermore, the OODL is always one of the low-level layers. This may be associated with the fact that high-level layers are customized to extract features that are useful to separate the classes of input and not to separate the distribution of ID dataset. In other words, high-level features cover unique cases (related to classes), whereas the low-level features cover general cases (related to the distribution) \citep{Zintgraf2017}. Figure \ref{id_ood_umap_1} demonstrates this perspective using two-dimensional representations of extracted features for the penultimate and OODL of the ResNet model. As seen, the features extracted by the penultimate layer are helpful to separate the classes of input, whereas features of the OODL are useful to separate the distribution of the ID dataset.

\subsection{Input preprocessing}
As indicated in Section \ref{intoduction}, input preprocessing is a technique used by ODIN and related approaches to increase the predicted softmax class probability and thus improve the detection of ID inputs. In our approach, we do not use the output of softmax, but input preprocessing can increase feature values for ID inputs, so that the OSVM classifier can better detect ID inputs. Therefore, we can additionally exploit input preprocessing with our approach; however, Section \ref{Experimental_results} shows that our approach outperforms other approaches even without input preprocessing.

To implement input preprocessing, we first calculate the pre-processed sample $x^{\prime}$, for each input $x$ at the test time, by adding a small perturbation and then use the features of $x^{\prime}$ for detection. Input $x^{\prime}$ is obtained as follows
\begin{equation}
\label{equ:perturbation}
x^{\prime}=x-\varepsilon \cdot \mathrm{sign} \Big( - \nabla_{x} \mathrm{log}  \big(  \mathrm{max}_{i} \: Q_{i}  \left( x \right) \big) \Big)
\end{equation}
where $\varepsilon$ is the perturbation magnitude, and the perturbation is calculated by back-propagating the gradients of the predicted class probability with respect to input $x$.

\begin{table*}
  \tiny
  \caption{Comparison of our approach with the max-softmax, ODIN, and MD approaches for the MNIST datasets. Values are in percentages and $\downarrow$ indicates that lower values are better, while $\uparrow$ indicates that higher values are better. \textbf{Bold} indicates the best score.}
  \label{mnist_no_input_preprocessing}
  \centering
  \begin{tabular}{lllllll}
    \toprule
    \shortstack{ID\\model}     & OOD     & \shortstack{FPR at\\ 95\% TPR} $ \downarrow$ & \shortstack{Detection \\ error$\downarrow$}  & AUROC$\uparrow$  & \shortstack {AUPR\\Out$\uparrow$}  & \shortstack{AUPR\\In$\uparrow$} \\
    \cmidrule(l){3-7}
    \multicolumn{5}{r}{ max-softmax / ODIN / MD / ours}                   \\
    \midrule
\multirow{4}{*}{\shortstack{MNIST\\CUSTOM-CNN}}
&\multicolumn{1}{l}{F-MNIST}
&\multicolumn{1}{l}{6.77/5.75/34.51/\textbf{0.42}}
&\multicolumn{1}{l}{5.82/5.32/19.74/\textbf{2.67}}
&\multicolumn{1}{l}{98.06/98.77/86.86/\textbf{99.08}}
&\multicolumn{1}{l}{98.44/98.85/90.03/\textbf{98.84}}
&\multicolumn{1}{l}{97.55/98.71/82.09/\textbf{99.29}}
\\
&\multicolumn{1}{l}{Omniglot}
&\multicolumn{1}{l}{6.92/4.79/10.15/\textbf{0.0}}
&\multicolumn{1}{l}{5.82/4.53/7.55/\textbf{0.0}}
&\multicolumn{1}{l}{97.64/98.8/97.35/\textbf{100.0}}
&\multicolumn{1}{l}{98.3/99.04/97.91/\textbf{100.0}}
&\multicolumn{1}{l}{96.51/98.5/96.71/\textbf{100.0}}
\\
&\multicolumn{1}{l}{Gaussian}
&\multicolumn{1}{l}{1.74/0.13/26.72/\textbf{0.0}}
&\multicolumn{1}{l}{1.22/0.15/14.76/\textbf{0.0}}
&\multicolumn{1}{l}{98.91/99.96/77.67/\textbf{100.0}}
&\multicolumn{1}{l}{99.38/99.97/87.46/\textbf{100.0}}
&\multicolumn{1}{l}{96.86/99.91/62.24/\textbf{100.0}}
\\
&\multicolumn{1}{l}{Uniform}
&\multicolumn{1}{l}{3.76/0.8/30.68/\textbf{0.0}}
&\multicolumn{1}{l}{2.87/1.04/17.5/\textbf{0.0}}
&\multicolumn{1}{l}{98.11/99.74/78.24/\textbf{100.0}}
&\multicolumn{1}{l}{98.84/99.81/87.0/\textbf{100.0}}
&\multicolumn{1}{l}{95.74/99.58/63.29/\textbf{100.0}}
\\

    \bottomrule
  \end{tabular}
\end{table*}

\section{Experiments} \label{Experimental_results}
We apply our approach to several ID and OOD datasets under different learning models. We compare our approach with max-softmax, ODIN, and MD (over the logits layer with input preprocessing) approaches. As already discussed, we focus on approaches that do not require training on OOD samples or retraining the classifier. Using OOD samples gives access to extra knowledge and retraining a classifier may be very costly. However, to respect the established practice in the literature, we include input preprocessing (with access to OOD) as an optional step in our comparison. Due to the page limitation, other comparisons including uncertainty-based approaches are in supplemental material. Our implementation is available online for reproducibility.

\subsection{ID datasets and models}
We evaluate our approach over several ID datasets, including MNIST, CIFAR-10, and CIFAR-100 \citep{cifar_2009}. MNIST is a dataset of handwritten digits and has $60,000$ images in the training set and $10,000$ images on the test set. It includes $28\times28$ grayscale images. We trained a custom convolutional neural network (CNN) for the MNIST dataset with two convolutional layers and two fully connected layers. The model has an accuracy of $99.22\%$ in the test set. The CIFAR-10 dataset has $50,000$ and $10,000$ images for training and testing, respectively. It includes $32\times32$ colored images. We trained two models based on VGG-16 by \citet{Simonyan_vgg16_2014} and ResNet (i.e., ResNet44 v1) by ~\citet{resnet_He_2015} for CIFAR-10, achieving an accuracy of $93.56\%$ and $92.01 \%$, respectively. CIFAR-100 is similar to CIFAR-10, but it has $100$ classes. We also trained VGG-16 and ResNet models for CIFAR-100, with an accuracy of $70.48\%$ and $69.17\%$, respectively. Test sets of all ID datasets are used to compute metrics defined in Section \ref{evaluation-metrics}.

\subsection{OOD datasets}
We consider several OOD datasets for our evaluation, including synthetic ones, and use their test sets for computing the metrics. Moreover, we always keep the size of ID and OOD inputs the same during evaluation (by randomly selecting data from the larger dataset to match the number of instances in the smaller dataset). The following are the OOD datasets used for our experiments.

\begin{itemize}
     \item \textbf{Fashion-MNIST (F-MNIST)} is similar to the MNIST dataset, but it includes Zalando's article images \citep{FMNIST_2017}.
     \item \textbf{Omniglot} contains different handwritten characters from 50 different alphabets. The images have been downsampled to $28 \times28$ images \citep{omniglot_Lake1332}. 
     \item \textbf{TinyImageNet} consists of a subset of ImageNet images \citep{imagenet_2009} and covers $200$ different classes. We downsampled images to $32\times32$.
     \item \textbf{LSUN} includes $32\times32$ downsampled images from the Large-scale Scene UNderstanding dataset \citep{LSUN_2015}. 
     \item \textbf{iSUN} includes $32\times32$ downsampled images of iSUN images \cite{iSUN_2015}.
     \item \textbf{SVHN} includes real-world images similar to the MNIST dataset \citep{svhn_2011}.
     \item \textbf{Gaussian noise} includes random normal noise with $\mu = 0.5$ and $\sigma = 1$, clipped to [0, 1].
     \item \textbf{Uniform noise} includes random uniform noise between [0, 1].
    \end{itemize}

\subsection{Evaluation metrics} \label{evaluation-metrics}
There are different metrics to evaluate the performance of OOD detection approaches. We adopted the following metrics \citep{Hendrycks-baseline-2017,Liang-ODIN-2018}.
\begin{itemize}
 \item FPR at $95\%$ TPR is the probability of an out-of-distribution (i.e., negative) input being misclassified as in-distribution (i.e., positive) input when the true positive rate (TPR) is as high as $95\%$. True positive rate is calculated by $\textit{TPR} = \textit{TP} / (\textit{TP}+\textit{FN})$, where TP and FN denote true positives and false negatives, respectively. The false positive rate (FPR) is computed by $\textit{FPR} = \textit{FP} / (\textit{FP}+\textit{TN})$, where FP and TN denote false positives and true negatives, respectively.
 \item Detection error calculates the misclassification probability when TPR is $95\%$. It is equal to $0.5*(1 - \textit{TPR}) + 0.5*\textit{FPR}$, where we assume that both positive and negative examples have an equal probability of appearing in the evaluation test.
 \item AUROC is the Area Under the Receiver Operating Characteristic curve. It is interpreted as the probability that a positive example is assigned a higher detection score than a negative example. An ideal OOD detector expects an AUROC score of $100\%$.
 \item AUPR is the Area under the Precision-Recall curve. It is a graph reflecting precision equal to $\textit{TP}/(\textit{TP}+\textit{FP})$ and recall equal to $\textit{TP}/(\textit{TP}+\textit{FN})$ against each other. The metric AUPR-In and AUPR-Out represent the area under the precision-recall curve where in-distribution or out-of-distribution images are specified as positives, respectively.
\end{itemize}

\subsection{OSVM and hyper parameters}
To train OSVM classifiers for detection and finding the OODL, we used the \textit{rbf} kernel and training error $\textit{nu}=0.001$. The \textit{rbf} kernel gave us the best results in comparison to other kernels such as \textit{linear} or \textit{poly}. We used temperature scale $T=1000$ for the ODIN approach since it was deemed as the optimal value based in the original paper \citep{Liang-ODIN-2018}. The perturbation magnitude $\varepsilon$ for our approach and ODIN was optimized to minimize FPR at $95\%$ TPR by having access to randomly selected $20\%$ of OOD datasets. We used F-MNIST and TinyImageNet datasets to find the OODLs. The OODL was fixed for every ID dataset and its associated model. The OODL for the MNIST dataset was the second convolutional layer. The OODLs for VGG-16 trained on CIFAR-10 and CIFAR-100 were the second convolutional layer and the first max-polling layer, respectively. The OODL for ResNet trained on CIFAR-10 and CIFAR-100 were the thirteenth and ninth residual layers, respectively. Details of parameters for training models are in supplemental material. 
\subsection{Detection results}
Table \ref{mnist_no_input_preprocessing} shows the comparison of our approach with others for MNIST (a low-dimensional dataset). Our approach gives better results across all the metrics defined in Section \ref{evaluation-metrics}. The comparison of our approach \textit{with} input preprocessing with CIFAR-10 and CIDAR-100 are listed in Table \ref{cifarx_odin_vs_ours}.a. Our approach also gives better results for these two datasets. Furthermore, our approach is capable of fully detecting noise for both low and high dimensional data under different models. Table \ref{cifarx_odin_vs_ours}.b compares our approach \textit{without} input preprocessing with others (\textit{with} input preprocessing). As seen, it outperforms other approaches except in the cases of TinyImageNet for CIFAR-100. Our approach still has better AUROC, but the detection error and FPR at $95\%$ TPR are slightly larger than ODIN's. Interestingly, the MD approach is worse than max-softmax in some cases. Such a result has also been reported by \citet{ren_ratio_2019} in their Table 3 results.


\begin{table*}[t!]
  \tiny
  \caption{Comparison of our approach with the max-softmax, ODIN, and MD approaches for the CIFAR-10 and CIFAR-100 datasets. a) Our approach \textit{with} input preprocessing. b) Our approach \textit{without} input preprocessing. $\downarrow$ indicates that lower values are better, whereas $\uparrow$ indicates that higher values are better. \textbf{Bold} indicates the best score.}
  \label{cifarx_odin_vs_ours}
  \centering
  \subcaption*{a)}
    \begin{tabular}{lllllll}
    \toprule
    \shortstack{ID\\model}     & OOD     & \shortstack{FPR at\\ 95\% TPR} $ \downarrow$ & \shortstack{Detection \\ error$\downarrow$}  & AUROC$\uparrow$  & \shortstack {AUPR\\Out$\uparrow$}  & \shortstack{AUPR\\In$\uparrow$} \\
    \cmidrule(l){3-7}
    \multicolumn{5}{r}{ max-softmax / ODIN / MD / ours}                   \\
    \midrule
\multirow{4}{*}{\shortstack{CIFAR-10\\VGG16}}
&\multicolumn{1}{l}{TinyImagenet}
&\multicolumn{1}{l}{35.53/32.96/31.76/\textbf{15.89}}
&\multicolumn{1}{l}{20.26/18.98/18.37/\textbf{10.44}}
&\multicolumn{1}{l}{87.64/91.36/90.89/\textbf{97.06}}
&\multicolumn{1}{l}{89.89/91.88/91.7/\textbf{96.82}}
&\multicolumn{1}{l}{83.6/90.34/89.41/\textbf{97.34}}
\\
&\multicolumn{1}{l}{LSUN}
&\multicolumn{1}{l}{27.4/16.14/17.46/\textbf{7.03}}
&\multicolumn{1}{l}{16.19/10.56/11.22/\textbf{6.01}}
&\multicolumn{1}{l}{90.17/96.01/95.52/\textbf{98.54}}
&\multicolumn{1}{l}{92.62/96.52/96.07/\textbf{98.54}}
&\multicolumn{1}{l}{86.37/95.33/94.72/\textbf{98.58}}
\\
&\multicolumn{1}{l}{iSUN}
&\multicolumn{1}{l}{28.66/17.98/19.28/\textbf{8.4}}
&\multicolumn{1}{l}{16.82/11.48/12.13/\textbf{6.7}}
&\multicolumn{1}{l}{89.58/95.49/94.93/\textbf{98.41}}
&\multicolumn{1}{l}{92.2/96.07/95.67/\textbf{98.43}}
&\multicolumn{1}{l}{85.49/94.81/93.99/\textbf{98.47}}
\\
&\multicolumn{1}{l}{SVHN}
&\multicolumn{1}{l}{28.35/23.66/29.58/\textbf{8.46}}
&\multicolumn{1}{l}{16.65/14.33/17.29/\textbf{6.7}}
&\multicolumn{1}{l}{89.05/92.27/91.57/\textbf{97.32}}
&\multicolumn{1}{l}{92.02/93.92/92.44/\textbf{97.95}}
&\multicolumn{1}{l}{83.46/89.15/89.63/\textbf{95.64}}
\\
&\multicolumn{1}{l}{Gaussian}
&\multicolumn{1}{l}{20.62/12.87/3.11/\textbf{0.0}}
&\multicolumn{1}{l}{12.32/8.82/3.02/\textbf{0.0}}
&\multicolumn{1}{l}{86.76/95.25/99.09/\textbf{100.0}}
&\multicolumn{1}{l}{92.09/96.7/99.35/\textbf{100.0}}
&\multicolumn{1}{l}{73.63/91.29/98.6/\textbf{100.0}}
\\
&\multicolumn{1}{l}{Uniform}
&\multicolumn{1}{l}{25.19/40.31/6.27/\textbf{0.0}}
&\multicolumn{1}{l}{14.72/22.65/4.62/\textbf{0.0}}
&\multicolumn{1}{l}{82.22/80.33/97.18/\textbf{100.0}}
&\multicolumn{1}{l}{89.56/86.27/98.18/\textbf{100.0}}
&\multicolumn{1}{l}{67.35/68.42/94.09/\textbf{100.0}}
\\
\midrule
\multirow{4}{*}{\shortstack{CIFAR-10\\ResNet-V1-44}}
&\multicolumn{1}{l}{TinyImagenet}
&\multicolumn{1}{l}{36.15/19.88/68.38/\textbf{5.47}}
&\multicolumn{1}{l}{20.56/12.42/36.66/\textbf{5.21}}
&\multicolumn{1}{l}{87.69/95.2/73.82/\textbf{98.71}}
&\multicolumn{1}{l}{90.16/95.68/76.05/\textbf{98.46}}
&\multicolumn{1}{l}{84.03/94.67/68.86/\textbf{98.91}}
\\
&\multicolumn{1}{l}{LSUN}
&\multicolumn{1}{l}{27.83/8.72/63.53/\textbf{2.47}}
&\multicolumn{1}{l}{16.4/6.85/34.26/\textbf{3.68}}
&\multicolumn{1}{l}{90.32/97.99/73.48/\textbf{99.16}}
&\multicolumn{1}{l}{92.6/98.27/77.49/\textbf{99.01}}
&\multicolumn{1}{l}{87.11/97.66/65.53/\textbf{99.3}}
\\
&\multicolumn{1}{l}{iSUN}
&\multicolumn{1}{l}{29.98/9.95/66.17/\textbf{5.54}}
&\multicolumn{1}{l}{17.45/7.45/35.58/\textbf{5.27}}
&\multicolumn{1}{l}{89.63/97.73/72.83/\textbf{98.85}}
&\multicolumn{1}{l}{92.01/98.03/76.19/\textbf{98.72}}
&\multicolumn{1}{l}{86.37/97.42/66.09/\textbf{99.0}}
\\
&\multicolumn{1}{l}{SVHN}
&\multicolumn{1}{l}{21.17/9.75/16.91/\textbf{4.44}}
&\multicolumn{1}{l}{13.07/7.36/10.94/\textbf{4.66}}
&\multicolumn{1}{l}{92.98/97.63/96.63/\textbf{98.68}}
&\multicolumn{1}{l}{94.54/97.97/96.67/\textbf{98.78}}
&\multicolumn{1}{l}{90.74/96.75/96.64/\textbf{98.12}}
\\
&\multicolumn{1}{l}{Gaussian}
&\multicolumn{1}{l}{38.17/50.06/0.0/\textbf{0.0}}
&\multicolumn{1}{l}{21.55/27.52/0.0/\textbf{0.0}}
&\multicolumn{1}{l}{84.22/73.68/100.0/\textbf{100.0}}
&\multicolumn{1}{l}{88.3/81.34/100.0/\textbf{100.0}}
&\multicolumn{1}{l}{75.84/61.14/100.0/\textbf{100.0}}
\\
&\multicolumn{1}{l}{Uniform}
&\multicolumn{1}{l}{18.66/9.79/0.0/\textbf{0.0}}
&\multicolumn{1}{l}{11.82/7.24/0.01/\textbf{0.0}}
&\multicolumn{1}{l}{92.67/96.36/100.0/\textbf{100.0}}
&\multicolumn{1}{l}{94.8/97.49/100.0/\textbf{100.0}}
&\multicolumn{1}{l}{88.94/92.35/100.0/\textbf{100.0}}
\\
\midrule
\multirow{4}{*}{\shortstack{CIFAR-100\\VGG16}}
&\multicolumn{1}{l}{TinyImagenet}
&\multicolumn{1}{l}{63.56/50.66/57.7/\textbf{21.9}}
&\multicolumn{1}{l}{34.24/27.81/31.34/\textbf{13.44}}
&\multicolumn{1}{l}{74.04/84.2/80.73/\textbf{95.35}}
&\multicolumn{1}{l}{77.29/85.83/82.36/\textbf{95.64}}
&\multicolumn{1}{l}{69.3/80.96/77.29/\textbf{95.24}}
\\
&\multicolumn{1}{l}{LSUN}
&\multicolumn{1}{l}{61.73/45.7/53.29/\textbf{19.53}}
&\multicolumn{1}{l}{33.36/25.32/29.11/\textbf{12.26}}
&\multicolumn{1}{l}{73.73/84.84/82.85/\textbf{96.4}}
&\multicolumn{1}{l}{77.99/87.16/84.43/\textbf{96.43}}
&\multicolumn{1}{l}{68.12/80.65/79.75/\textbf{96.5}}
\\
&\multicolumn{1}{l}{iSUN}
&\multicolumn{1}{l}{64.42/49.21/56.78/\textbf{22.7}}
&\multicolumn{1}{l}{34.71/27.1/30.89/\textbf{13.85}}
&\multicolumn{1}{l}{72.48/83.97/81.82/\textbf{95.39}}
&\multicolumn{1}{l}{76.68/86.08/82.94/\textbf{95.6}}
&\multicolumn{1}{l}{66.89/79.84/78.61/\textbf{95.44}}
\\
&\multicolumn{1}{l}{SVHN}
&\multicolumn{1}{l}{65.49/45.32/49.46/\textbf{22.23}}
&\multicolumn{1}{l}{35.24/25.15/27.16/\textbf{13.6}}
&\multicolumn{1}{l}{71.89/84.71/82.19/\textbf{92.5}}
&\multicolumn{1}{l}{76.29/87.24/85.31/\textbf{94.23}}
&\multicolumn{1}{l}{67.58/79.06/76.64/\textbf{87.44}}
\\
&\multicolumn{1}{l}{Gaussian}
&\multicolumn{1}{l}{94.89/61.31/26.16/\textbf{0.0}}
&\multicolumn{1}{l}{48.52/32.9/15.07/\textbf{0.0}}
&\multicolumn{1}{l}{12.49/48.84/82.6/\textbf{100.0}}
&\multicolumn{1}{l}{36.69/67.58/89.53/\textbf{100.0}}
&\multicolumn{1}{l}{32.92/44.25/68.31/\textbf{100.0}}
\\
&\multicolumn{1}{l}{Uniform}
&\multicolumn{1}{l}{96.53/68.54/45.4/\textbf{0.0}}
&\multicolumn{1}{l}{48.96/36.3/25.08/\textbf{0.0}}
&\multicolumn{1}{l}{8.09/42.19/65.67/\textbf{100.0}}
&\multicolumn{1}{l}{34.45/62.47/78.88/\textbf{100.0}}
&\multicolumn{1}{l}{32.09/41.57/53.21/\textbf{100.0}}
\\
\midrule
\multirow{4}{*}{\shortstack{CIFAR-100\\ResNet-V1-44}}
&\multicolumn{1}{l}{TinyImagenet}
&\multicolumn{1}{l}{63.39/44.09/79.7/\textbf{13.65}}
&\multicolumn{1}{l}{34.18/24.52/42.35/\textbf{9.32}}
&\multicolumn{1}{l}{75.1/87.91/66.81/\textbf{97.69}}
&\multicolumn{1}{l}{78.35/88.82/68.17/\textbf{97.56}}
&\multicolumn{1}{l}{71.11/86.52/62.0/\textbf{97.92}}
\\
&\multicolumn{1}{l}{LSUN}
&\multicolumn{1}{l}{61.72/31.25/78.07/\textbf{18.38}}
&\multicolumn{1}{l}{33.36/18.11/41.53/\textbf{11.68}}
&\multicolumn{1}{l}{75.36/92.15/66.55/\textbf{97.01}}
&\multicolumn{1}{l}{78.9/93.0/69.18/\textbf{96.81}}
&\multicolumn{1}{l}{71.11/91.13/60.61/\textbf{97.35}}
\\
&\multicolumn{1}{l}{iSUN}
&\multicolumn{1}{l}{61.67/35.6/85.99/\textbf{20.59}}
&\multicolumn{1}{l}{33.33/20.3/45.49/\textbf{12.79}}
&\multicolumn{1}{l}{74.62/90.74/61.57/\textbf{96.62}}
&\multicolumn{1}{l}{78.46/91.81/62.23/\textbf{96.46}}
&\multicolumn{1}{l}{70.23/89.43/56.9/\textbf{96.99}}
\\
&\multicolumn{1}{l}{SVHN}
&\multicolumn{1}{l}{58.13/19.94/12.76/\textbf{11.35}}
&\multicolumn{1}{l}{31.55/12.46/8.88/\textbf{8.17}}
&\multicolumn{1}{l}{78.63/92.54/97.63/\textbf{97.25}}
&\multicolumn{1}{l}{81.72/94.39/97.57/\textbf{97.59}}
&\multicolumn{1}{l}{75.75/87.2/97.68/\textbf{95.98}}
\\
&\multicolumn{1}{l}{Gaussian}
&\multicolumn{1}{l}{68.46/89.33/0.0/\textbf{0.0}}
&\multicolumn{1}{l}{36.61/46.16/0.0/\textbf{0.0}}
&\multicolumn{1}{l}{51.19/18.84/100.0/\textbf{100.0}}
&\multicolumn{1}{l}{65.82/43.07/100.0/\textbf{100.0}}
&\multicolumn{1}{l}{45.41/34.49/100.0/\textbf{100.0}}
\\
&\multicolumn{1}{l}{Uniform}
&\multicolumn{1}{l}{47.66/59.41/0.04/\textbf{0.0}}
&\multicolumn{1}{l}{26.32/31.96/0.14/\textbf{0.0}}
&\multicolumn{1}{l}{72.59/53.11/99.99/\textbf{100.0}}
&\multicolumn{1}{l}{81.13/69.97/99.99/\textbf{100.0}}
&\multicolumn{1}{l}{60.56/46.16/99.97/\textbf{100.0}}
\\

    \bottomrule
  \end{tabular}
  \subcaption*{b)}
  \begin{tabular}{lllllll}
    \toprule
    \shortstack{ID\\model}     & OOD     & \shortstack{FPR at\\ 95\% TPR} $ \downarrow$ & \shortstack{Detection \\ error$\downarrow$}  & AUROC$\uparrow$  & \shortstack {AUPR\\Out$\uparrow$}  & \shortstack{AUPR\\In$\uparrow$} \\
    \cmidrule(l){3-7}
    \multicolumn{5}{r}{ max-softmax / ODIN / MD / ours without preprocessing}                   \\
    \midrule
\multirow{4}{*}{\shortstack{CIFAR-10\\VGG16}}
&\multicolumn{1}{l}{TinyImagenet}
&\multicolumn{1}{l}{35.53/32.96/31.76/\textbf{25.6}}
&\multicolumn{1}{l}{20.26/18.98/18.37/\textbf{15.29}}
&\multicolumn{1}{l}{87.64/91.36/90.89/\textbf{95.91}}
&\multicolumn{1}{l}{89.89/91.88/91.7/\textbf{95.21}}
&\multicolumn{1}{l}{83.6/90.34/89.41/\textbf{96.52}}
\\
&\multicolumn{1}{l}{LSUN}
&\multicolumn{1}{l}{27.4/16.14/17.46/\textbf{8.75}}
&\multicolumn{1}{l}{16.19/10.56/11.22/\textbf{6.87}}
&\multicolumn{1}{l}{90.17/96.01/95.52/\textbf{98.21}}
&\multicolumn{1}{l}{92.62/96.52/96.07/\textbf{98.1}}
&\multicolumn{1}{l}{86.37/95.33/94.72/\textbf{98.36}}
\\
&\multicolumn{1}{l}{iSUN}
&\multicolumn{1}{l}{28.66/17.98/19.28/\textbf{10.66}}
&\multicolumn{1}{l}{16.82/11.48/12.13/\textbf{7.82}}
&\multicolumn{1}{l}{89.58/95.49/94.93/\textbf{98.04}}
&\multicolumn{1}{l}{92.2/96.07/95.67/\textbf{97.93}}
&\multicolumn{1}{l}{85.49/94.81/93.99/\textbf{98.21}}
\\
&\multicolumn{1}{l}{SVHN}
&\multicolumn{1}{l}{28.35/23.66/29.58/\textbf{8.46}}
&\multicolumn{1}{l}{16.65/14.33/17.29/\textbf{6.7}}
&\multicolumn{1}{l}{89.05/92.27/91.57/\textbf{97.32}}
&\multicolumn{1}{l}{92.02/93.92/92.44/\textbf{97.95}}
&\multicolumn{1}{l}{83.46/89.15/89.63/\textbf{95.64}}
\\
&\multicolumn{1}{l}{Gaussian}
&\multicolumn{1}{l}{20.62/12.87/3.11/\textbf{0.0}}
&\multicolumn{1}{l}{12.32/8.82/3.02/\textbf{0.0}}
&\multicolumn{1}{l}{86.76/95.25/99.09/\textbf{100.0}}
&\multicolumn{1}{l}{92.09/96.7/99.35/\textbf{100.0}}
&\multicolumn{1}{l}{73.63/91.29/98.6/\textbf{100.0}}
\\
&\multicolumn{1}{l}{Uniform}
&\multicolumn{1}{l}{25.19/40.31/6.27/\textbf{0.0}}
&\multicolumn{1}{l}{14.72/22.65/4.62/\textbf{0.0}}
&\multicolumn{1}{l}{82.22/80.33/97.18/\textbf{100.0}}
&\multicolumn{1}{l}{89.56/86.27/98.18/\textbf{100.0}}
&\multicolumn{1}{l}{67.35/68.42/94.09/\textbf{100.0}}
\\
\midrule
\multirow{4}{*}{\shortstack{CIFAR-10\\ResNet-V1-44}}
&\multicolumn{1}{l}{TinyImagenet}
&\multicolumn{1}{l}{36.15/19.88/68.38/\textbf{7.93}}
&\multicolumn{1}{l}{20.56/12.42/36.66/\textbf{6.45}}
&\multicolumn{1}{l}{87.69/95.2/73.82/\textbf{98.32}}
&\multicolumn{1}{l}{90.16/95.68/76.05/\textbf{97.83}}
&\multicolumn{1}{l}{84.03/94.67/68.86/\textbf{98.65}}
\\
&\multicolumn{1}{l}{LSUN}
&\multicolumn{1}{l}{27.83/8.72/63.53/\textbf{2.49}}
&\multicolumn{1}{l}{16.4/6.85/34.26/\textbf{3.73}}
&\multicolumn{1}{l}{90.32/97.99/73.48/\textbf{99.14}}
&\multicolumn{1}{l}{92.6/98.27/77.49/\textbf{98.99}}
&\multicolumn{1}{l}{87.11/97.66/65.53/\textbf{99.29}}
\\
&\multicolumn{1}{l}{iSUN}
&\multicolumn{1}{l}{29.98/9.95/66.17/\textbf{7.38}}
&\multicolumn{1}{l}{17.45/7.45/35.58/\textbf{6.19}}
&\multicolumn{1}{l}{89.63/97.73/72.83/\textbf{98.57}}
&\multicolumn{1}{l}{92.01/98.03/76.19/\textbf{98.31}}
&\multicolumn{1}{l}{86.37/97.42/66.09/\textbf{98.81}}
\\
&\multicolumn{1}{l}{SVHN}
&\multicolumn{1}{l}{21.17/9.75/16.91/\textbf{4.44}}
&\multicolumn{1}{l}{13.07/7.36/10.94/\textbf{4.66}}
&\multicolumn{1}{l}{92.98/97.63/96.63/\textbf{98.68}}
&\multicolumn{1}{l}{94.54/97.97/96.67/\textbf{98.78}}
&\multicolumn{1}{l}{90.74/96.75/96.64/\textbf{98.12}}
\\
&\multicolumn{1}{l}{Gaussian}
&\multicolumn{1}{l}{38.17/50.06/0.0/\textbf{0.0}}
&\multicolumn{1}{l}{21.55/27.52/0.0/\textbf{0.0}}
&\multicolumn{1}{l}{84.22/73.68/100.0/\textbf{100.0}}
&\multicolumn{1}{l}{88.3/81.34/100.0/\textbf{100.0}}
&\multicolumn{1}{l}{75.84/61.14/100.0/\textbf{100.0}}
\\
&\multicolumn{1}{l}{Uniform}
&\multicolumn{1}{l}{18.66/9.79/0.0/\textbf{0.0}}
&\multicolumn{1}{l}{11.82/7.24/0.01/\textbf{0.0}}
&\multicolumn{1}{l}{92.67/96.36/100.0/\textbf{100.0}}
&\multicolumn{1}{l}{94.8/97.49/100.0/\textbf{100.0}}
&\multicolumn{1}{l}{88.94/92.35/100.0/\textbf{100.0}}
\\
\midrule
\multirow{4}{*}{\shortstack{CIFAR-100\\VGG16}}
&\multicolumn{1}{l}{TinyImagenet}
&\multicolumn{1}{l}{63.56/\textbf{50.66}/57.7/51.87}
&\multicolumn{1}{l}{34.24/\textbf{27.81}/31.34/28.43}
&\multicolumn{1}{l}{74.04/84.2/80.73/\textbf{91.14}}
&\multicolumn{1}{l}{77.29/85.83/82.36/\textbf{89.0}}
&\multicolumn{1}{l}{69.3/80.96/77.29/\textbf{92.73}}
\\
&\multicolumn{1}{l}{LSUN}
&\multicolumn{1}{l}{61.73/45.7/53.29/\textbf{28.14}}
&\multicolumn{1}{l}{33.36/25.32/29.11/\textbf{16.55}}
&\multicolumn{1}{l}{73.73/84.84/82.85/\textbf{95.36}}
&\multicolumn{1}{l}{77.99/87.16/84.43/\textbf{94.62}}
&\multicolumn{1}{l}{68.12/80.65/79.75/\textbf{96.04}}
\\
&\multicolumn{1}{l}{iSUN}
&\multicolumn{1}{l}{64.42/49.21/56.78/\textbf{31.27}}
&\multicolumn{1}{l}{34.71/27.1/30.89/\textbf{18.13}}
&\multicolumn{1}{l}{72.48/83.97/81.82/\textbf{94.66}}
&\multicolumn{1}{l}{76.68/86.08/82.94/\textbf{93.98}}
&\multicolumn{1}{l}{66.89/79.84/78.61/\textbf{95.42}}
\\
&\multicolumn{1}{l}{SVHN}
&\multicolumn{1}{l}{65.49/45.32/49.46/\textbf{22.23}}
&\multicolumn{1}{l}{35.24/25.15/27.16/\textbf{13.6}}
&\multicolumn{1}{l}{71.89/84.71/82.19/\textbf{92.5}}
&\multicolumn{1}{l}{76.29/87.24/85.31/\textbf{94.23}}
&\multicolumn{1}{l}{67.58/79.06/76.64/\textbf{87.44}}
\\
&\multicolumn{1}{l}{Gaussian}
&\multicolumn{1}{l}{94.89/61.31/26.16/\textbf{0.0}}
&\multicolumn{1}{l}{48.52/32.9/15.07/\textbf{0.0}}
&\multicolumn{1}{l}{12.49/48.84/82.6/\textbf{100.0}}
&\multicolumn{1}{l}{36.69/67.58/89.53/\textbf{100.0}}
&\multicolumn{1}{l}{32.92/44.25/68.31/\textbf{100.0}}
\\
&\multicolumn{1}{l}{Uniform}
&\multicolumn{1}{l}{96.53/68.54/45.4/\textbf{0.0}}
&\multicolumn{1}{l}{48.96/36.3/25.08/\textbf{0.0}}
&\multicolumn{1}{l}{8.09/42.19/65.67/\textbf{100.0}}
&\multicolumn{1}{l}{34.45/62.47/78.88/\textbf{100.0}}
&\multicolumn{1}{l}{32.09/41.57/53.21/\textbf{100.0}}
\\
\midrule
\multirow{4}{*}{\shortstack{CIFAR-100\\ResNet-V1-44}}
&\multicolumn{1}{l}{TinyImagenet}
&\multicolumn{1}{l}{63.39/44.09/79.7/\textbf{20.45}}
&\multicolumn{1}{l}{34.18/24.52/42.35/\textbf{12.7}}
&\multicolumn{1}{l}{75.1/87.91/66.81/\textbf{96.77}}
&\multicolumn{1}{l}{78.35/88.82/68.17/\textbf{96.32}}
&\multicolumn{1}{l}{71.11/86.52/62.0/\textbf{97.24}}
\\
&\multicolumn{1}{l}{LSUN}
&\multicolumn{1}{l}{61.72/31.25/78.07/\textbf{19.68}}
&\multicolumn{1}{l}{33.36/18.11/41.53/\textbf{12.33}}
&\multicolumn{1}{l}{75.36/92.15/66.55/\textbf{96.81}}
&\multicolumn{1}{l}{78.9/93.0/69.18/\textbf{96.55}}
&\multicolumn{1}{l}{71.11/91.13/60.61/\textbf{97.21}}
\\
&\multicolumn{1}{l}{iSUN}
&\multicolumn{1}{l}{61.67/35.6/85.99/\textbf{21.92}}
&\multicolumn{1}{l}{33.33/20.3/45.49/\textbf{13.46}}
&\multicolumn{1}{l}{74.62/90.74/61.57/\textbf{96.37}}
&\multicolumn{1}{l}{78.46/91.81/62.23/\textbf{96.14}}
&\multicolumn{1}{l}{70.23/89.43/56.9/\textbf{96.81}}
\\
&\multicolumn{1}{l}{SVHN}
&\multicolumn{1}{l}{58.13/19.94/12.76/\textbf{12.15}}
&\multicolumn{1}{l}{31.55/12.46/8.88/\textbf{8.55}}
&\multicolumn{1}{l}{78.63/92.54/97.63/\textbf{96.95}}
&\multicolumn{1}{l}{81.72/94.39/97.57/\textbf{97.4}}
&\multicolumn{1}{l}{75.75/87.2/97.68/\textbf{95.2}}
\\
&\multicolumn{1}{l}{Gaussian}
&\multicolumn{1}{l}{68.46/89.33/0.0/\textbf{0.0}}
&\multicolumn{1}{l}{36.61/46.16/0.0/\textbf{0.0}}
&\multicolumn{1}{l}{51.19/18.84/100.0/\textbf{100.0}}
&\multicolumn{1}{l}{65.82/43.07/100.0/\textbf{100.0}}
&\multicolumn{1}{l}{45.41/34.49/100.0/\textbf{100.0}}
\\
&\multicolumn{1}{l}{Uniform}
&\multicolumn{1}{l}{47.66/59.41/0.04/\textbf{0.0}}
&\multicolumn{1}{l}{26.32/31.96/0.14/\textbf{0.0}}
&\multicolumn{1}{l}{72.59/53.11/99.99/\textbf{100.0}}
&\multicolumn{1}{l}{81.13/69.97/99.99/\textbf{100.0}}
&\multicolumn{1}{l}{60.56/46.16/99.97/\textbf{100.0}}
\\
    \bottomrule
  \end{tabular}
\end{table*}

\section{Conclusion}
Detecting OOD inputs of DNNs is an important concern for the application of DNNs in safety-related domains, such as autonomous driving and medical diagnostics. Most of the current OOD detection approaches rely on features extracted by the penultimate layers of DNNs. Such a layer is trained to extract features that are relevant to separate the classes of input. In contrast, the proposed OOD detection approach relies on the key empirical finding that one of the early layers of a deep classifier, which we refer to as the optimal OOD discernment layer, provides a suitable latent space in which the distributions of ID and OOD datasets are well separated. This result allows us to train one-class classifier to detect OOD inputs, without requiring access to OOD samples at training. However, when such samples are available the approach can be extended by input preprocessing to improve the detection. We experimentally evaluated our approach on several low- and high dimensional datasets and deep models. The results show substantial improvement over the baseline and the state-of-the-art approaches for multiple metrics. As part of our future work, we plan to explore using a dedicated loss function to improve features extracted by the OODL for both OOD detection and input class separation.

\bibliographystyle{aaai}
\bibliography{bibfile}

\section{Supplemental material}
\subsection{Other comparisons}
We report comparison of our approach \textit{(without preprocessing)} with different uncertainty estimation metrics including entropy \cite{Shannon1948}, margin\cite{Scheffer_margin_2001}, MC-dropout \cite{Gal-dropout-2016}, and mutual information between predictions and model posterior \cite{Gal_bald_2017}. To calculate uncertainty metrics based on MC-dropout we run 100 times the model for each input while all dropout layers are enabled. The results are shown in Tables \ref{mnist_entropy} - \ref{cifar_mutual_info}. 

We also compare our results with ODIN and MD (Mahalanobis Distance) \textit{without preprocessing}. Their results are also listed in tables \ref{mnist_odin}-\ref{cifar_md}. Our approach still outperforms these approaches. As expected, the performance of these approaches is reduced because they do not have access to OOD samples anymore.

\subsection{Hyper parameters}
We used three different models for the experiments. Their hyper parameters are listed in table \ref{model_hyper_parameters}. The best perturbation magnitude for each dataset was obtained from the following vector [0.0, 0.0005, 0.001, 0.0015, 0.002, 0.0025, 0.005, 0.01, 0.05, 0.1, 0.15, 0.2]. This vector is fixed for all approaches including ours.

\begin{table*} [h]
\caption{Comparison between our approach (without preprocessing) and entropy for the MNIST datasets. Values are in percentages and $\downarrow$ indicates that lower values are better, while $\uparrow$ indicates that higher values are better. \textbf{Bold} indicates the best score.}
  \label{mnist_entropy}
  \centering
  \begin{tabular}{lllllll}
    \toprule
    \shortstack{ID\\model}     & OOD     & \shortstack{FPR at\\ 95\% TPR} $ \downarrow$ & \shortstack{Detection \\ error$\downarrow$}  & AUROC$\uparrow$  & \shortstack {AUPR\\Out$\uparrow$}  & \shortstack{AUPR\\In$\uparrow$} \\
    \cmidrule(l){3-7}
    \multicolumn{5}{r}{ entropy / ours without preprocessing}                   \\
    \midrule
\multirow{4}{*}{\shortstack{MNIST\\CUSTOM-CNN}}
&\multicolumn{1}{l}{F-MNIST}
&\multicolumn{1}{l}{6.66/\textbf{0.42}}
&\multicolumn{1}{l}{5.73/\textbf{2.71}}
&\multicolumn{1}{l}{98.29/\textbf{99.04}}
&\multicolumn{1}{l}{98.57/\textbf{98.78}}
&\multicolumn{1}{l}{98.02/\textbf{99.27}}
\\
&\multicolumn{1}{l}{Omniglot}
&\multicolumn{1}{l}{6.67/\textbf{0.0}}
&\multicolumn{1}{l}{5.69/\textbf{0.0}}
&\multicolumn{1}{l}{97.94/\textbf{100.0}}
&\multicolumn{1}{l}{98.45/\textbf{100.0}}
&\multicolumn{1}{l}{97.25/\textbf{100.0}}
\\
&\multicolumn{1}{l}{Gaussian}
&\multicolumn{1}{l}{0.84/\textbf{0.0}}
&\multicolumn{1}{l}{0.99/\textbf{0.0}}
&\multicolumn{1}{l}{99.72/\textbf{100.0}}
&\multicolumn{1}{l}{99.81/\textbf{100.0}}
&\multicolumn{1}{l}{99.54/\textbf{100.0}}
\\
&\multicolumn{1}{l}{Uniform}
&\multicolumn{1}{l}{3.38/\textbf{0.0}}
&\multicolumn{1}{l}{2.72/\textbf{0.0}}
&\multicolumn{1}{l}{98.69/\textbf{100.0}}
&\multicolumn{1}{l}{99.11/\textbf{100.0}}
&\multicolumn{1}{l}{97.94/\textbf{100.0}}
\\
\\
  \bottomrule
  \end{tabular}
\end{table*}
\begin{table*}
  \caption{Comparison between our approach (without preprocessing) and margin for the MNIST datasets. Values are in percentages and $\downarrow$ indicates that lower values are better, while $\uparrow$ indicates that higher values are better. \textbf{Bold} indicates the best score.}
  \label{mnist_margin}
  \centering
  \begin{tabular}{lllllll}
    \toprule
    \shortstack{ID\\model}     & OOD     & \shortstack{FPR at\\ 95\% TPR} $ \downarrow$ & \shortstack{Detection \\ error$\downarrow$}  & AUROC$\uparrow$  & \shortstack {AUPR\\Out$\uparrow$}  & \shortstack{AUPR\\In$\uparrow$} \\
    \cmidrule(l){3-7}
    \multicolumn{5}{r}{ margin / ours without preprocessing}                   \\
    \midrule
\multirow{4}{*}{\shortstack{MNIST\\CUSTOM-CNN}}
&\multicolumn{1}{l}{F-MNIST}
&\multicolumn{1}{l}{7.06/\textbf{0.42}}
&\multicolumn{1}{l}{5.92/\textbf{2.71}}
&\multicolumn{1}{l}{97.88/\textbf{99.04}}
&\multicolumn{1}{l}{98.35/\textbf{98.78}}
&\multicolumn{1}{l}{96.94/\textbf{99.27}}
\\
&\multicolumn{1}{l}{Omniglot}
&\multicolumn{1}{l}{7.33/\textbf{0.0}}
&\multicolumn{1}{l}{6.02/\textbf{0.0}}
&\multicolumn{1}{l}{97.43/\textbf{100.0}}
&\multicolumn{1}{l}{98.18/\textbf{100.0}}
&\multicolumn{1}{l}{95.8/\textbf{100.0}}
\\
&\multicolumn{1}{l}{Gaussian}
&\multicolumn{1}{l}{2.11/\textbf{0.0}}
&\multicolumn{1}{l}{1.52/\textbf{0.0}}
&\multicolumn{1}{l}{98.48/\textbf{100.0}}
&\multicolumn{1}{l}{99.17/\textbf{100.0}}
&\multicolumn{1}{l}{94.64/\textbf{100.0}}
\\
&\multicolumn{1}{l}{Uniform}
&\multicolumn{1}{l}{4.2/\textbf{0.0}}
&\multicolumn{1}{l}{3.04/\textbf{0.0}}
&\multicolumn{1}{l}{97.78/\textbf{100.0}}
&\multicolumn{1}{l}{98.67/\textbf{100.0}}
&\multicolumn{1}{l}{93.98/\textbf{100.0}}
\\
  \bottomrule
  \end{tabular}
\end{table*}
\begin{table*}
  \caption{Comparison between our approach (without preprocessing) and MC-dropout for the MNIST datasets. Values are in percentages and $\downarrow$ indicates that lower values are better, while $\uparrow$ indicates that higher values are better. \textbf{Bold} indicates the best score.}
  \label{mnist_dropout}
  \centering
  \begin{tabular}{lllllll}
    \toprule
    \shortstack{ID\\model}     & OOD     & \shortstack{FPR at\\ 95\% TPR} $ \downarrow$ & \shortstack{Detection \\ error$\downarrow$}  & AUROC$\uparrow$  & \shortstack {AUPR\\Out$\uparrow$}  & \shortstack{AUPR\\In$\uparrow$} \\
    \cmidrule(l){3-7}
    \multicolumn{5}{r}{ MC-dropout / ours without preprocessing}                   \\
    \midrule
\multirow{4}{*}{\shortstack{MNIST\\CUSTOM-CNN}}
&\multicolumn{1}{l}{F-MNIST}
&\multicolumn{1}{l}{6.59/\textbf{0.42}}
&\multicolumn{1}{l}{5.53/\textbf{2.71}}
&\multicolumn{1}{l}{98.19/\textbf{99.04}}
&\multicolumn{1}{l}{98.58/\textbf{98.78}}
&\multicolumn{1}{l}{97.71/\textbf{99.27}}
\\
&\multicolumn{1}{l}{Omniglot}
&\multicolumn{1}{l}{6.54/\textbf{0.0}}
&\multicolumn{1}{l}{5.3/\textbf{0.0}}
&\multicolumn{1}{l}{97.78/\textbf{100.0}}
&\multicolumn{1}{l}{98.41/\textbf{100.0}}
&\multicolumn{1}{l}{96.72/\textbf{100.0}}
\\
&\multicolumn{1}{l}{Gaussian}
&\multicolumn{1}{l}{2.3/\textbf{0.0}}
&\multicolumn{1}{l}{1.66/\textbf{0.0}}
&\multicolumn{1}{l}{98.71/\textbf{100.0}}
&\multicolumn{1}{l}{99.24/\textbf{100.0}}
&\multicolumn{1}{l}{96.44/\textbf{100.0}}
\\
&\multicolumn{1}{l}{Uniform}
&\multicolumn{1}{l}{5.03/\textbf{0.0}}
&\multicolumn{1}{l}{3.71/\textbf{0.0}}
&\multicolumn{1}{l}{97.45/\textbf{100.0}}
&\multicolumn{1}{l}{98.43/\textbf{100.0}}
&\multicolumn{1}{l}{94.43/\textbf{100.0}}
\\
  \bottomrule
  \end{tabular}
\end{table*}
\begin{table*}
  \caption{Comparison between our approach (without preprocessing) and mutual information for the MNIST datasets. Values are in percentages and $\downarrow$ indicates that lower values are better, while $\uparrow$ indicates that higher values are better. \textbf{Bold} indicates the best score.}
  \label{mnist_mutual_info}
  \centering
  \begin{tabular}{lllllll}
    \toprule
    \shortstack{ID\\model}     & OOD     & \shortstack{FPR at\\ 95\% TPR} $ \downarrow$ & \shortstack{Detection \\ error$\downarrow$}  & AUROC$\uparrow$  & \shortstack {AUPR\\Out$\uparrow$}  & \shortstack{AUPR\\In$\uparrow$} \\
    \cmidrule(l){3-7}
    \multicolumn{5}{r}{ mutual information / ours without preprocessing}                   \\
    \midrule
\multirow{4}{*}{\shortstack{MNIST\\CUSTOM-CNN}}
&\multicolumn{1}{l}{F-MNIST}
&\multicolumn{1}{l}{7.27/\textbf{0.42}}
&\multicolumn{1}{l}{5.71/\textbf{2.71}}
&\multicolumn{1}{l}{97.53/\textbf{99.04}}
&\multicolumn{1}{l}{98.21/\textbf{98.78}}
&\multicolumn{1}{l}{96.24/\textbf{99.27}}
\\
&\multicolumn{1}{l}{Omniglot}
&\multicolumn{1}{l}{6.6/\textbf{0.0}}
&\multicolumn{1}{l}{5.42/\textbf{0.0}}
&\multicolumn{1}{l}{97.6/\textbf{100.0}}
&\multicolumn{1}{l}{98.33/\textbf{100.0}}
&\multicolumn{1}{l}{95.75/\textbf{100.0}}
\\
&\multicolumn{1}{l}{Gaussian}
&\multicolumn{1}{l}{4.29/\textbf{0.0}}
&\multicolumn{1}{l}{2.71/\textbf{0.0}}
&\multicolumn{1}{l}{96.72/\textbf{100.0}}
&\multicolumn{1}{l}{98.22/\textbf{100.0}}
&\multicolumn{1}{l}{89.69/\textbf{100.0}}
\\
&\multicolumn{1}{l}{Uniform}
&\multicolumn{1}{l}{7.1/\textbf{0.0}}
&\multicolumn{1}{l}{4.79/\textbf{0.0}}
&\multicolumn{1}{l}{95.19/\textbf{100.0}}
&\multicolumn{1}{l}{97.3/\textbf{100.0}}
&\multicolumn{1}{l}{86.65/\textbf{100.0}}
\\
  \bottomrule
  \end{tabular}
\end{table*}

\begin{table*}
  \caption{Comparison between our approach (without preprocessing) and entropy for the CIFAR-10 and CIFAR-100 datasets. $\downarrow$ indicates that lower values are better, whereas $\uparrow$ indicates that higher values are better. \textbf{Bold} indicates the best score.}
  \label{cifar_entropy}
  \centering
  \begin{tabular}{lllllll}
    \toprule
    \shortstack{ID\\model}     & OOD     & \shortstack{FPR at\\ 95\% TPR} $ \downarrow$ & \shortstack{Detection \\ error$\downarrow$}  & AUROC$\uparrow$  & \shortstack {AUPR\\Out$\uparrow$}  & \shortstack{AUPR\\In$\uparrow$} \\
    \cmidrule(l){3-7}
    \multicolumn{5}{r}{ entropy / ours without preprocessing}                   \\
    \midrule
\multirow{4}{*}{\shortstack{CIFAR10\\VGG16}}
&\multicolumn{1}{l}{TinyImagenet}
&\multicolumn{1}{l}{35.44/\textbf{25.6}}
&\multicolumn{1}{l}{20.22/\textbf{15.29}}
&\multicolumn{1}{l}{88.11/\textbf{95.91}}
&\multicolumn{1}{l}{90.11/\textbf{95.21}}
&\multicolumn{1}{l}{85.16/\textbf{96.52}}
\\
&\multicolumn{1}{l}{LSUN}
&\multicolumn{1}{l}{26.87/\textbf{8.75}}
&\multicolumn{1}{l}{15.92/\textbf{6.87}}
&\multicolumn{1}{l}{90.8/\textbf{98.21}}
&\multicolumn{1}{l}{92.92/\textbf{98.1}}
&\multicolumn{1}{l}{88.2/\textbf{98.36}}
\\
&\multicolumn{1}{l}{iSUN}
&\multicolumn{1}{l}{28.19/\textbf{10.66}}
&\multicolumn{1}{l}{16.56/\textbf{7.82}}
&\multicolumn{1}{l}{90.18/\textbf{98.04}}
&\multicolumn{1}{l}{92.49/\textbf{97.93}}
&\multicolumn{1}{l}{87.33/\textbf{98.21}}
\\
&\multicolumn{1}{l}{SVHN}
&\multicolumn{1}{l}{27.86/\textbf{8.46}}
&\multicolumn{1}{l}{16.42/\textbf{6.7}}
&\multicolumn{1}{l}{89.42/\textbf{97.32}}
&\multicolumn{1}{l}{92.22/\textbf{97.95}}
&\multicolumn{1}{l}{84.6/\textbf{95.64}}
\\
&\multicolumn{1}{l}{Gaussian}
&\multicolumn{1}{l}{20.43/\textbf{0.0}}
&\multicolumn{1}{l}{12.24/\textbf{0.0}}
&\multicolumn{1}{l}{86.96/\textbf{100.0}}
&\multicolumn{1}{l}{92.19/\textbf{100.0}}
&\multicolumn{1}{l}{74.06/\textbf{100.0}}
\\
&\multicolumn{1}{l}{Uniform}
&\multicolumn{1}{l}{25.15/\textbf{0.0}}
&\multicolumn{1}{l}{14.74/\textbf{0.0}}
&\multicolumn{1}{l}{82.25/\textbf{100.0}}
&\multicolumn{1}{l}{89.56/\textbf{100.0}}
&\multicolumn{1}{l}{67.39/\textbf{100.0}}
\\
\midrule
\multirow{4}{*}{\shortstack{CIFAR10\\ResNet-V1-44}}
&\multicolumn{1}{l}{TinyImagenet}
&\multicolumn{1}{l}{36.14/\textbf{7.93}}
&\multicolumn{1}{l}{20.55/\textbf{6.45}}
&\multicolumn{1}{l}{88.28/\textbf{98.32}}
&\multicolumn{1}{l}{90.41/\textbf{97.83}}
&\multicolumn{1}{l}{85.55/\textbf{98.65}}
\\
&\multicolumn{1}{l}{LSUN}
&\multicolumn{1}{l}{27.58/\textbf{2.49}}
&\multicolumn{1}{l}{16.28/\textbf{3.73}}
&\multicolumn{1}{l}{91.09/\textbf{99.14}}
&\multicolumn{1}{l}{92.97/\textbf{98.99}}
&\multicolumn{1}{l}{88.88/\textbf{99.29}}
\\
&\multicolumn{1}{l}{iSUN}
&\multicolumn{1}{l}{29.62/\textbf{7.38}}
&\multicolumn{1}{l}{17.31/\textbf{6.19}}
&\multicolumn{1}{l}{90.4/\textbf{98.57}}
&\multicolumn{1}{l}{92.37/\textbf{98.31}}
&\multicolumn{1}{l}{88.18/\textbf{98.81}}
\\
&\multicolumn{1}{l}{SVHN}
&\multicolumn{1}{l}{21.0/\textbf{4.44}}
&\multicolumn{1}{l}{13.0/\textbf{4.66}}
&\multicolumn{1}{l}{93.77/\textbf{98.68}}
&\multicolumn{1}{l}{94.93/\textbf{98.78}}
&\multicolumn{1}{l}{92.34/\textbf{98.12}}
\\
&\multicolumn{1}{l}{Gaussian}
&\multicolumn{1}{l}{38.29/\textbf{0.0}}
&\multicolumn{1}{l}{21.64/\textbf{0.0}}
&\multicolumn{1}{l}{83.47/\textbf{100.0}}
&\multicolumn{1}{l}{88.02/\textbf{100.0}}
&\multicolumn{1}{l}{72.79/\textbf{100.0}}
\\
&\multicolumn{1}{l}{Uniform}
&\multicolumn{1}{l}{18.39/\textbf{0.0}}
&\multicolumn{1}{l}{11.63/\textbf{0.0}}
&\multicolumn{1}{l}{93.15/\textbf{100.0}}
&\multicolumn{1}{l}{95.05/\textbf{100.0}}
&\multicolumn{1}{l}{89.01/\textbf{100.0}}
\\
\midrule
\multirow{4}{*}{\shortstack{CIFAR100\\VGG16}}
&\multicolumn{1}{l}{TinyImagenet}
&\multicolumn{1}{l}{63.07/\textbf{51.87}}
&\multicolumn{1}{l}{34.03/\textbf{28.43}}
&\multicolumn{1}{l}{75.36/\textbf{91.14}}
&\multicolumn{1}{l}{78.01/\textbf{89.0}}
&\multicolumn{1}{l}{71.35/\textbf{92.73}}
\\
&\multicolumn{1}{l}{LSUN}
&\multicolumn{1}{l}{61.31/\textbf{28.14}}
&\multicolumn{1}{l}{33.11/\textbf{16.55}}
&\multicolumn{1}{l}{74.84/\textbf{95.36}}
&\multicolumn{1}{l}{78.63/\textbf{94.62}}
&\multicolumn{1}{l}{69.7/\textbf{96.04}}
\\
&\multicolumn{1}{l}{iSUN}
&\multicolumn{1}{l}{63.93/\textbf{31.27}}
&\multicolumn{1}{l}{34.46/\textbf{18.13}}
&\multicolumn{1}{l}{73.62/\textbf{94.66}}
&\multicolumn{1}{l}{77.33/\textbf{93.98}}
&\multicolumn{1}{l}{68.48/\textbf{95.42}}
\\
&\multicolumn{1}{l}{SVHN}
&\multicolumn{1}{l}{64.82/\textbf{22.23}}
&\multicolumn{1}{l}{34.9/\textbf{13.6}}
&\multicolumn{1}{l}{73.32/\textbf{92.5}}
&\multicolumn{1}{l}{77.1/\textbf{94.23}}
&\multicolumn{1}{l}{69.83/\textbf{87.44}}
\\
&\multicolumn{1}{l}{Gaussian}
&\multicolumn{1}{l}{94.82/\textbf{0.0}}
&\multicolumn{1}{l}{48.5/\textbf{0.0}}
&\multicolumn{1}{l}{12.55/\textbf{100.0}}
&\multicolumn{1}{l}{36.75/\textbf{100.0}}
&\multicolumn{1}{l}{32.93/\textbf{100.0}}
\\
&\multicolumn{1}{l}{Uniform}
&\multicolumn{1}{l}{96.49/\textbf{0.0}}
&\multicolumn{1}{l}{48.94/\textbf{0.0}}
&\multicolumn{1}{l}{8.23/\textbf{100.0}}
&\multicolumn{1}{l}{34.52/\textbf{100.0}}
&\multicolumn{1}{l}{32.11/\textbf{100.0}}
\\
\midrule
\multirow{4}{*}{\shortstack{CIFAR100\\ResNet-V1-44}}
&\multicolumn{1}{l}{TinyImagenet}
&\multicolumn{1}{l}{62.59/\textbf{20.45}}
&\multicolumn{1}{l}{33.76/\textbf{12.7}}
&\multicolumn{1}{l}{77.38/\textbf{96.77}}
&\multicolumn{1}{l}{79.72/\textbf{96.32}}
&\multicolumn{1}{l}{74.01/\textbf{97.24}}
\\
&\multicolumn{1}{l}{LSUN}
&\multicolumn{1}{l}{61.12/\textbf{19.68}}
&\multicolumn{1}{l}{33.05/\textbf{12.33}}
&\multicolumn{1}{l}{78.0/\textbf{96.81}}
&\multicolumn{1}{l}{80.45/\textbf{96.55}}
&\multicolumn{1}{l}{74.76/\textbf{97.21}}
\\
&\multicolumn{1}{l}{iSUN}
&\multicolumn{1}{l}{61.12/\textbf{21.92}}
&\multicolumn{1}{l}{33.03/\textbf{13.46}}
&\multicolumn{1}{l}{76.95/\textbf{96.37}}
&\multicolumn{1}{l}{79.82/\textbf{96.14}}
&\multicolumn{1}{l}{73.29/\textbf{96.81}}
\\
&\multicolumn{1}{l}{SVHN}
&\multicolumn{1}{l}{57.17/\textbf{12.15}}
&\multicolumn{1}{l}{31.06/\textbf{8.55}}
&\multicolumn{1}{l}{81.87/\textbf{96.95}}
&\multicolumn{1}{l}{83.74/\textbf{97.4}}
&\multicolumn{1}{l}{79.93/\textbf{95.2}}
\\
&\multicolumn{1}{l}{Gaussian}
&\multicolumn{1}{l}{68.39/\textbf{0.0}}
&\multicolumn{1}{l}{36.58/\textbf{0.0}}
&\multicolumn{1}{l}{50.01/\textbf{100.0}}
&\multicolumn{1}{l}{65.5/\textbf{100.0}}
&\multicolumn{1}{l}{44.76/\textbf{100.0}}
\\
&\multicolumn{1}{l}{Uniform}
&\multicolumn{1}{l}{47.66/\textbf{0.0}}
&\multicolumn{1}{l}{26.32/\textbf{0.0}}
&\multicolumn{1}{l}{71.05/\textbf{100.0}}
&\multicolumn{1}{l}{80.73/\textbf{100.0}}
&\multicolumn{1}{l}{57.96/\textbf{100.0}}
\\

  \bottomrule
  \end{tabular}
\end{table*}

\begin{table*}
  \caption{Comparison between our approach (without preprocessing) and margin for the CIFAR-10 and CIFAR-100 datasets. $\downarrow$ indicates that lower values are better, whereas $\uparrow$ indicates that higher values are better. \textbf{Bold} indicates the best score.}
  \label{cifar_margin}
  \centering
  \begin{tabular}{lllllll}
    \toprule
    \shortstack{ID\\model}     & OOD     & \shortstack{FPR at\\ 95\% TPR} $ \downarrow$ & \shortstack{Detection \\ error$\downarrow$}  & AUROC$\uparrow$  & \shortstack {AUPR\\Out$\uparrow$}  & \shortstack{AUPR\\In$\uparrow$} \\
    \cmidrule(l){3-7}
    \multicolumn{5}{r}{ margin / ours without preprocessing}                   \\
    \midrule
\multirow{4}{*}{\shortstack{CIFAR10\\VGG16}}
&\multicolumn{1}{l}{TinyImagenet}
&\multicolumn{1}{l}{36.17/\textbf{25.6}}
&\multicolumn{1}{l}{20.54/\textbf{15.29}}
&\multicolumn{1}{l}{87.26/\textbf{95.91}}
&\multicolumn{1}{l}{89.63/\textbf{95.21}}
&\multicolumn{1}{l}{82.51/\textbf{96.52}}
\\
&\multicolumn{1}{l}{LSUN}
&\multicolumn{1}{l}{28.73/\textbf{8.75}}
&\multicolumn{1}{l}{16.8/\textbf{6.87}}
&\multicolumn{1}{l}{89.71/\textbf{98.21}}
&\multicolumn{1}{l}{92.3/\textbf{98.1}}
&\multicolumn{1}{l}{85.02/\textbf{98.36}}
\\
&\multicolumn{1}{l}{iSUN}
&\multicolumn{1}{l}{29.94/\textbf{10.66}}
&\multicolumn{1}{l}{17.46/\textbf{7.82}}
&\multicolumn{1}{l}{89.12/\textbf{98.04}}
&\multicolumn{1}{l}{91.88/\textbf{97.93}}
&\multicolumn{1}{l}{84.1/\textbf{98.21}}
\\
&\multicolumn{1}{l}{SVHN}
&\multicolumn{1}{l}{29.33/\textbf{8.46}}
&\multicolumn{1}{l}{17.14/\textbf{6.7}}
&\multicolumn{1}{l}{88.75/\textbf{97.32}}
&\multicolumn{1}{l}{91.77/\textbf{97.95}}
&\multicolumn{1}{l}{82.97/\textbf{95.64}}
\\
&\multicolumn{1}{l}{Gaussian}
&\multicolumn{1}{l}{21.03/\textbf{0.0}}
&\multicolumn{1}{l}{12.54/\textbf{0.0}}
&\multicolumn{1}{l}{86.49/\textbf{100.0}}
&\multicolumn{1}{l}{91.91/\textbf{100.0}}
&\multicolumn{1}{l}{73.3/\textbf{100.0}}
\\
&\multicolumn{1}{l}{Uniform}
&\multicolumn{1}{l}{25.4/\textbf{0.0}}
&\multicolumn{1}{l}{14.79/\textbf{0.0}}
&\multicolumn{1}{l}{82.08/\textbf{100.0}}
&\multicolumn{1}{l}{89.47/\textbf{100.0}}
&\multicolumn{1}{l}{67.19/\textbf{100.0}}
\\
\midrule
\multirow{4}{*}{\shortstack{CIFAR10\\ResNet-V1-44}}
&\multicolumn{1}{l}{TinyImagenet}
&\multicolumn{1}{l}{36.37/\textbf{7.93}}
&\multicolumn{1}{l}{20.67/\textbf{6.45}}
&\multicolumn{1}{l}{87.3/\textbf{98.32}}
&\multicolumn{1}{l}{89.94/\textbf{97.83}}
&\multicolumn{1}{l}{82.56/\textbf{98.65}}
\\
&\multicolumn{1}{l}{LSUN}
&\multicolumn{1}{l}{28.37/\textbf{2.49}}
&\multicolumn{1}{l}{16.68/\textbf{3.73}}
&\multicolumn{1}{l}{89.83/\textbf{99.14}}
&\multicolumn{1}{l}{92.35/\textbf{98.99}}
&\multicolumn{1}{l}{85.42/\textbf{99.29}}
\\
&\multicolumn{1}{l}{iSUN}
&\multicolumn{1}{l}{30.39/\textbf{7.38}}
&\multicolumn{1}{l}{17.69/\textbf{6.19}}
&\multicolumn{1}{l}{89.13/\textbf{98.57}}
&\multicolumn{1}{l}{91.75/\textbf{98.31}}
&\multicolumn{1}{l}{84.59/\textbf{98.81}}
\\
&\multicolumn{1}{l}{SVHN}
&\multicolumn{1}{l}{21.27/\textbf{4.44}}
&\multicolumn{1}{l}{13.12/\textbf{4.66}}
&\multicolumn{1}{l}{92.51/\textbf{98.68}}
&\multicolumn{1}{l}{94.31/\textbf{98.78}}
&\multicolumn{1}{l}{89.09/\textbf{98.12}}
\\
&\multicolumn{1}{l}{Gaussian}
&\multicolumn{1}{l}{37.84/\textbf{0.0}}
&\multicolumn{1}{l}{21.42/\textbf{0.0}}
&\multicolumn{1}{l}{84.6/\textbf{100.0}}
&\multicolumn{1}{l}{88.5/\textbf{100.0}}
&\multicolumn{1}{l}{77.67/\textbf{100.0}}
\\
&\multicolumn{1}{l}{Uniform}
&\multicolumn{1}{l}{19.1/\textbf{0.0}}
&\multicolumn{1}{l}{12.03/\textbf{0.0}}
&\multicolumn{1}{l}{92.34/\textbf{100.0}}
&\multicolumn{1}{l}{94.62/\textbf{100.0}}
&\multicolumn{1}{l}{88.01/\textbf{100.0}}
\\
\midrule
\multirow{4}{*}{\shortstack{CIFAR100\\VGG16}}
&\multicolumn{1}{l}{TinyImagenet}
&\multicolumn{1}{l}{64.32/\textbf{51.87}}
&\multicolumn{1}{l}{34.64/\textbf{28.43}}
&\multicolumn{1}{l}{73.21/\textbf{91.14}}
&\multicolumn{1}{l}{76.74/\textbf{89.0}}
&\multicolumn{1}{l}{67.55/\textbf{92.73}}
\\
&\multicolumn{1}{l}{LSUN}
&\multicolumn{1}{l}{62.63/\textbf{28.14}}
&\multicolumn{1}{l}{33.8/\textbf{16.55}}
&\multicolumn{1}{l}{73.08/\textbf{95.36}}
&\multicolumn{1}{l}{77.53/\textbf{94.62}}
&\multicolumn{1}{l}{66.99/\textbf{96.04}}
\\
&\multicolumn{1}{l}{iSUN}
&\multicolumn{1}{l}{65.05/\textbf{31.27}}
&\multicolumn{1}{l}{35.03/\textbf{18.13}}
&\multicolumn{1}{l}{71.78/\textbf{94.66}}
&\multicolumn{1}{l}{76.17/\textbf{93.98}}
&\multicolumn{1}{l}{65.8/\textbf{95.42}}
\\
&\multicolumn{1}{l}{SVHN}
&\multicolumn{1}{l}{66.64/\textbf{22.23}}
&\multicolumn{1}{l}{35.81/\textbf{13.6}}
&\multicolumn{1}{l}{70.95/\textbf{92.5}}
&\multicolumn{1}{l}{75.6/\textbf{94.23}}
&\multicolumn{1}{l}{65.51/\textbf{87.44}}
\\
&\multicolumn{1}{l}{Gaussian}
&\multicolumn{1}{l}{95.49/\textbf{0.0}}
&\multicolumn{1}{l}{48.69/\textbf{0.0}}
&\multicolumn{1}{l}{12.09/\textbf{100.0}}
&\multicolumn{1}{l}{36.14/\textbf{100.0}}
&\multicolumn{1}{l}{32.81/\textbf{100.0}}
\\
&\multicolumn{1}{l}{Uniform}
&\multicolumn{1}{l}{96.79/\textbf{0.0}}
&\multicolumn{1}{l}{49.05/\textbf{0.0}}
&\multicolumn{1}{l}{7.56/\textbf{100.0}}
&\multicolumn{1}{l}{34.12/\textbf{100.0}}
&\multicolumn{1}{l}{31.98/\textbf{100.0}}
\\
\midrule
\multirow{4}{*}{\shortstack{CIFAR100\\ResNet-V1-44}}
&\multicolumn{1}{l}{TinyImagenet}
&\multicolumn{1}{l}{63.65/\textbf{20.45}}
&\multicolumn{1}{l}{34.32/\textbf{12.7}}
&\multicolumn{1}{l}{73.48/\textbf{96.77}}
&\multicolumn{1}{l}{77.47/\textbf{96.32}}
&\multicolumn{1}{l}{67.86/\textbf{97.24}}
\\
&\multicolumn{1}{l}{LSUN}
&\multicolumn{1}{l}{62.37/\textbf{19.68}}
&\multicolumn{1}{l}{33.67/\textbf{12.33}}
&\multicolumn{1}{l}{73.59/\textbf{96.81}}
&\multicolumn{1}{l}{77.92/\textbf{96.55}}
&\multicolumn{1}{l}{67.52/\textbf{97.21}}
\\
&\multicolumn{1}{l}{iSUN}
&\multicolumn{1}{l}{62.31/\textbf{21.92}}
&\multicolumn{1}{l}{33.65/\textbf{13.46}}
&\multicolumn{1}{l}{73.08/\textbf{96.37}}
&\multicolumn{1}{l}{77.62/\textbf{96.14}}
&\multicolumn{1}{l}{67.07/\textbf{96.81}}
\\
&\multicolumn{1}{l}{SVHN}
&\multicolumn{1}{l}{58.79/\textbf{12.15}}
&\multicolumn{1}{l}{31.88/\textbf{8.55}}
&\multicolumn{1}{l}{76.25/\textbf{96.95}}
&\multicolumn{1}{l}{80.44/\textbf{97.4}}
&\multicolumn{1}{l}{70.57/\textbf{95.2}}
\\
&\multicolumn{1}{l}{Gaussian}
&\multicolumn{1}{l}{68.77/\textbf{0.0}}
&\multicolumn{1}{l}{36.76/\textbf{0.0}}
&\multicolumn{1}{l}{51.88/\textbf{100.0}}
&\multicolumn{1}{l}{65.89/\textbf{100.0}}
&\multicolumn{1}{l}{46.11/\textbf{100.0}}
\\
&\multicolumn{1}{l}{Uniform}
&\multicolumn{1}{l}{48.27/\textbf{0.0}}
&\multicolumn{1}{l}{26.54/\textbf{0.0}}
&\multicolumn{1}{l}{73.29/\textbf{100.0}}
&\multicolumn{1}{l}{81.27/\textbf{100.0}}
&\multicolumn{1}{l}{63.15/\textbf{100.0}}
\\

  \bottomrule
  \end{tabular}
\end{table*}

\begin{table*}
  \caption{Comparison between our approach (without preprocessing) and MC-dropout for the CIFAR-10 and CIFAR-100 datasets. $\downarrow$ indicates that lower values are better, whereas $\uparrow$ indicates that higher values are better. \textbf{Bold} indicates the best score. We do not report results for ResNet because it does not use dropout layers.}
  \label{cifar_dropout}
  \centering
  \begin{tabular}{lllllll}
    \toprule
    \shortstack{ID\\model}     & OOD     & \shortstack{FPR at\\ 95\% TPR} $ \downarrow$ & \shortstack{Detection \\ error$\downarrow$}  & AUROC$\uparrow$  & \shortstack {AUPR\\Out$\uparrow$}  & \shortstack{AUPR\\In$\uparrow$} \\
    \cmidrule(l){3-7}
    \multicolumn{5}{r}{ margin / ours without preprocessing}                   \\
    \midrule
\multirow{4}{*}{\shortstack{CIFAR10\\VGG16}}
&\multicolumn{1}{l}{TinyImagenet}
&\multicolumn{1}{l}{42.38/\textbf{25.6}}
&\multicolumn{1}{l}{23.62/\textbf{15.29}}
&\multicolumn{1}{l}{87.17/\textbf{95.91}}
&\multicolumn{1}{l}{83.77/\textbf{95.21}}
&\multicolumn{1}{l}{85.54/\textbf{96.52}}
\\
&\multicolumn{1}{l}{LSUN}
&\multicolumn{1}{l}{26.77/\textbf{8.75}}
&\multicolumn{1}{l}{15.87/\textbf{6.87}}
&\multicolumn{1}{l}{91.9/\textbf{98.21}}
&\multicolumn{1}{l}{92.24/\textbf{98.1}}
&\multicolumn{1}{l}{90.24/\textbf{98.36}}
\\
&\multicolumn{1}{l}{iSUN}
&\multicolumn{1}{l}{29.64/\textbf{10.66}}
&\multicolumn{1}{l}{17.31/\textbf{7.82}}
&\multicolumn{1}{l}{90.69/\textbf{98.04}}
&\multicolumn{1}{l}{90.8/\textbf{97.93}}
&\multicolumn{1}{l}{88.98/\textbf{98.21}}
\\
&\multicolumn{1}{l}{SVHN}
&\multicolumn{1}{l}{26.39/\textbf{8.46}}
&\multicolumn{1}{l}{15.69/\textbf{6.7}}
&\multicolumn{1}{l}{91.76/\textbf{97.32}}
&\multicolumn{1}{l}{92.65/\textbf{97.95}}
&\multicolumn{1}{l}{89.96/\textbf{95.64}}
\\
&\multicolumn{1}{l}{Gaussian}
&\multicolumn{1}{l}{6.49/\textbf{0.0}}
&\multicolumn{1}{l}{5.56/\textbf{0.0}}
&\multicolumn{1}{l}{98.75/\textbf{100.0}}
&\multicolumn{1}{l}{98.94/\textbf{100.0}}
&\multicolumn{1}{l}{98.59/\textbf{100.0}}
\\
&\multicolumn{1}{l}{Uniform}
&\multicolumn{1}{l}{6.36/\textbf{0.0}}
&\multicolumn{1}{l}{5.43/\textbf{0.0}}
&\multicolumn{1}{l}{98.74/\textbf{100.0}}
&\multicolumn{1}{l}{98.94/\textbf{100.0}}
&\multicolumn{1}{l}{98.58/\textbf{100.0}}
\\
\midrule
\multirow{4}{*}{\shortstack{CIFAR100\\VGG16}}
&\multicolumn{1}{l}{TinyImagenet}
&\multicolumn{1}{l}{61.02/\textbf{51.87}}
&\multicolumn{1}{l}{33.0/\textbf{28.43}}
&\multicolumn{1}{l}{78.28/\textbf{91.14}}
&\multicolumn{1}{l}{79.9/\textbf{89.0}}
&\multicolumn{1}{l}{74.57/\textbf{92.73}}
\\
&\multicolumn{1}{l}{LSUN}
&\multicolumn{1}{l}{45.08/\textbf{28.14}}
&\multicolumn{1}{l}{25.04/\textbf{16.55}}
&\multicolumn{1}{l}{86.26/\textbf{95.36}}
&\multicolumn{1}{l}{87.91/\textbf{94.62}}
&\multicolumn{1}{l}{84.19/\textbf{96.04}}
\\
&\multicolumn{1}{l}{iSUN}
&\multicolumn{1}{l}{55.92/\textbf{31.27}}
&\multicolumn{1}{l}{30.44/\textbf{18.13}}
&\multicolumn{1}{l}{82.63/\textbf{94.66}}
&\multicolumn{1}{l}{83.93/\textbf{93.98}}
&\multicolumn{1}{l}{80.58/\textbf{95.42}}
\\
&\multicolumn{1}{l}{SVHN}
&\multicolumn{1}{l}{41.88/\textbf{22.23}}
&\multicolumn{1}{l}{23.43/\textbf{13.6}}
&\multicolumn{1}{l}{87.47/\textbf{92.5}}
&\multicolumn{1}{l}{89.25/\textbf{94.23}}
&\multicolumn{1}{l}{85.27/\textbf{87.44}}
\\
&\multicolumn{1}{l}{Gaussian}
&\multicolumn{1}{l}{6.37/\textbf{0.0}}
&\multicolumn{1}{l}{5.67/\textbf{0.0}}
&\multicolumn{1}{l}{98.78/\textbf{100.0}}
&\multicolumn{1}{l}{98.89/\textbf{100.0}}
&\multicolumn{1}{l}{98.73/\textbf{100.0}}
\\
&\multicolumn{1}{l}{Uniform}
&\multicolumn{1}{l}{5.91/\textbf{0.0}}
&\multicolumn{1}{l}{5.43/\textbf{0.0}}
&\multicolumn{1}{l}{98.86/\textbf{100.0}}
&\multicolumn{1}{l}{98.97/\textbf{100.0}}
&\multicolumn{1}{l}{98.81/\textbf{100.0}}
\\

  \bottomrule
  \end{tabular}
\end{table*}
\begin{table*}
  \caption{Comparison between our approach (without preprocessing) and mutual information for the CIFAR-10 and CIFAR-100 datasets. $\downarrow$ indicates that lower values are better, whereas $\uparrow$ indicates that higher values are better. \textbf{Bold} indicates the best score. We do not report results for ResNet because it does not use dropout layers.}
  \label{cifar_mutual_info}
  \centering
  \begin{tabular}{lllllll}
    \toprule
    \shortstack{ID\\model}     & OOD     & \shortstack{FPR at\\ 95\% TPR} $ \downarrow$ & \shortstack{Detection \\ error$\downarrow$}  & AUROC$\uparrow$  & \shortstack {AUPR\\Out$\uparrow$}  & \shortstack{AUPR\\In$\uparrow$} \\
    \cmidrule(l){3-7}
    \multicolumn{5}{r}{ mutual information / ours without preprocessing}                   \\
    \midrule
\multirow{4}{*}{\shortstack{CIFAR10\\VGG16}}
&\multicolumn{1}{l}{TinyImagenet}
&\multicolumn{1}{l}{39.97/\textbf{25.6}}
&\multicolumn{1}{l}{22.48/\textbf{15.29}}
&\multicolumn{1}{l}{90.13/\textbf{95.91}}
&\multicolumn{1}{l}{86.51/\textbf{95.21}}
&\multicolumn{1}{l}{90.67/\textbf{96.52}}
\\
&\multicolumn{1}{l}{LSUN}
&\multicolumn{1}{l}{23.29/\textbf{8.75}}
&\multicolumn{1}{l}{14.14/\textbf{6.87}}
&\multicolumn{1}{l}{95.41/\textbf{98.21}}
&\multicolumn{1}{l}{94.93/\textbf{98.1}}
&\multicolumn{1}{l}{95.64/\textbf{98.36}}
\\
&\multicolumn{1}{l}{iSUN}
&\multicolumn{1}{l}{27.7/\textbf{10.66}}
&\multicolumn{1}{l}{16.34/\textbf{7.82}}
&\multicolumn{1}{l}{93.86/\textbf{98.04}}
&\multicolumn{1}{l}{93.31/\textbf{97.93}}
&\multicolumn{1}{l}{94.04/\textbf{98.21}}
\\
&\multicolumn{1}{l}{SVHN}
&\multicolumn{1}{l}{23.59/\textbf{8.46}}
&\multicolumn{1}{l}{14.28/\textbf{6.7}}
&\multicolumn{1}{l}{95.1/\textbf{97.32}}
&\multicolumn{1}{l}{95.36/\textbf{97.95}}
&\multicolumn{1}{l}{95.12/\textbf{95.64}}
\\
&\multicolumn{1}{l}{Gaussian}
&\multicolumn{1}{l}{0.04/\textbf{0.0}}
&\multicolumn{1}{l}{1.05/\textbf{0.0}}
&\multicolumn{1}{l}{99.94/\textbf{100.0}}
&\multicolumn{1}{l}{99.94/\textbf{100.0}}
&\multicolumn{1}{l}{99.95/\textbf{100.0}}
\\
&\multicolumn{1}{l}{Uniform}
&\multicolumn{1}{l}{0.03/\textbf{0.0}}
&\multicolumn{1}{l}{0.96/\textbf{0.0}}
&\multicolumn{1}{l}{99.95/\textbf{100.0}}
&\multicolumn{1}{l}{99.95/\textbf{100.0}}
&\multicolumn{1}{l}{99.95/\textbf{100.0}}
\\
\midrule
\multirow{4}{*}{\shortstack{CIFAR100\\VGG16}}
&\multicolumn{1}{l}{TinyImagenet}
&\multicolumn{1}{l}{58.42/\textbf{51.87}}
&\multicolumn{1}{l}{31.68/\textbf{28.43}}
&\multicolumn{1}{l}{83.61/\textbf{91.14}}
&\multicolumn{1}{l}{83.52/\textbf{89.0}}
&\multicolumn{1}{l}{83.38/\textbf{92.73}}
\\
&\multicolumn{1}{l}{LSUN}
&\multicolumn{1}{l}{34.39/\textbf{28.14}}
&\multicolumn{1}{l}{19.69/\textbf{16.55}}
&\multicolumn{1}{l}{93.19/\textbf{95.36}}
&\multicolumn{1}{l}{93.0/\textbf{94.62}}
&\multicolumn{1}{l}{93.7/\textbf{96.04}}
\\
&\multicolumn{1}{l}{iSUN}
&\multicolumn{1}{l}{50.75/\textbf{31.27}}
&\multicolumn{1}{l}{27.87/\textbf{18.13}}
&\multicolumn{1}{l}{89.0/\textbf{94.66}}
&\multicolumn{1}{l}{88.41/\textbf{93.98}}
&\multicolumn{1}{l}{89.79/\textbf{95.42}}
\\
&\multicolumn{1}{l}{SVHN}
&\multicolumn{1}{l}{27.96/\textbf{22.23}}
&\multicolumn{1}{l}{16.47/\textbf{13.6}}
&\multicolumn{1}{l}{94.7/\textbf{92.5}}
&\multicolumn{1}{l}{94.64/\textbf{94.23}}
&\multicolumn{1}{l}{95.05/\textbf{87.44}}
\\
&\multicolumn{1}{l}{Gaussian}
&\multicolumn{1}{l}{0.0/\textbf{0.0}}
&\multicolumn{1}{l}{0.29/\textbf{0.0}}
&\multicolumn{1}{l}{99.99/\textbf{100.0}}
&\multicolumn{1}{l}{99.99/\textbf{100.0}}
&\multicolumn{1}{l}{99.99/\textbf{100.0}}
\\
&\multicolumn{1}{l}{Uniform}
&\multicolumn{1}{l}{0.0/\textbf{0.0}}
&\multicolumn{1}{l}{0.26/\textbf{0.0}}
&\multicolumn{1}{l}{99.99/\textbf{100.0}}
&\multicolumn{1}{l}{99.99/\textbf{100.0}}
&\multicolumn{1}{l}{99.99/\textbf{100.0}}
\\

  \bottomrule
  \end{tabular}
\end{table*}

\begin{table*}
  \caption{Comparison between our approach (without preprocessing) and ODIN (without preprocessing) information for the MNIST datasets. Values are in percentages and $\downarrow$ indicates that lower values are better, while $\uparrow$ indicates that higher values are better. \textbf{Bold} indicates the best score.}
  \label{mnist_odin}
  \centering
  \begin{tabular}{lllllll}
    \toprule
    \shortstack{ID\\model}     & OOD     & \shortstack{FPR at\\ 95\% TPR} $ \downarrow$ & \shortstack{Detection \\ error$\downarrow$}  & AUROC$\uparrow$  & \shortstack {AUPR\\Out$\uparrow$}  & \shortstack{AUPR\\In$\uparrow$} \\
    \cmidrule(l){3-7}
    \multicolumn{5}{r}{ ODIN / ours (both without preprocessing)}                   \\
    \midrule
\multirow{4}{*}{\shortstack{MNIST\\CUSTOM-CNN}}
&\multicolumn{1}{l}{F-MNIST}
&\multicolumn{1}{l}{5.75/\textbf{0.42}}
&\multicolumn{1}{l}{5.32/\textbf{2.71}}
&\multicolumn{1}{l}{98.77/\textbf{99.04}}
&\multicolumn{1}{l}{98.85/\textbf{98.78}}
&\multicolumn{1}{l}{98.71/\textbf{99.27}}
\\
&\multicolumn{1}{l}{Omniglot}
&\multicolumn{1}{l}{5.33/\textbf{0.0}}
&\multicolumn{1}{l}{4.84/\textbf{0.0}}
&\multicolumn{1}{l}{98.63/\textbf{100.0}}
&\multicolumn{1}{l}{98.92/\textbf{100.0}}
&\multicolumn{1}{l}{98.23/\textbf{100.0}}
\\
&\multicolumn{1}{l}{Gaussian}
&\multicolumn{1}{l}{0.13/\textbf{0.0}}
&\multicolumn{1}{l}{0.15/\textbf{0.0}}
&\multicolumn{1}{l}{99.96/\textbf{100.0}}
&\multicolumn{1}{l}{99.97/\textbf{100.0}}
&\multicolumn{1}{l}{99.91/\textbf{100.0}}
\\
&\multicolumn{1}{l}{Uniform}
&\multicolumn{1}{l}{0.8/\textbf{0.0}}
&\multicolumn{1}{l}{1.04/\textbf{0.0}}
&\multicolumn{1}{l}{99.74/\textbf{100.0}}
&\multicolumn{1}{l}{99.81/\textbf{100.0}}
&\multicolumn{1}{l}{99.58/\textbf{100.0}}
\\
  \bottomrule
  \end{tabular}
\end{table*}

\begin{table*}
  \caption{Comparison between our approach (without preprocessing) and MD (without preprocessing) information for the MNIST datasets. Values are in percentages and $\downarrow$ indicates that lower values are better, while $\uparrow$ indicates that higher values are better. \textbf{Bold} indicates the best score.}
  \label{mnist_md}
  \centering
  \begin{tabular}{lllllll}
    \toprule
    \shortstack{ID\\model}     & OOD     & \shortstack{FPR at\\ 95\% TPR} $ \downarrow$ & \shortstack{Detection \\ error$\downarrow$}  & AUROC$\uparrow$  & \shortstack {AUPR\\Out$\uparrow$}  & \shortstack{AUPR\\In$\uparrow$} \\
    \cmidrule(l){3-7}
    \multicolumn{5}{r}{ MD / ours (both without preprocessing)}                   \\
    \midrule
\multirow{4}{*}{\shortstack{MNIST\\CUSTOM-CNN}}
&\multicolumn{1}{l}{F-MNIST}
&\multicolumn{1}{l}{46.43/\textbf{0.42}}
&\multicolumn{1}{l}{25.69/\textbf{2.71}}
&\multicolumn{1}{l}{82.71/\textbf{99.04}}
&\multicolumn{1}{l}{86.09/\textbf{98.78}}
&\multicolumn{1}{l}{78.39/\textbf{99.27}}
\\
&\multicolumn{1}{l}{Omniglot}
&\multicolumn{1}{l}{27.59/\textbf{0.0}}
&\multicolumn{1}{l}{16.27/\textbf{0.0}}
&\multicolumn{1}{l}{91.7/\textbf{100.0}}
&\multicolumn{1}{l}{93.42/\textbf{100.0}}
&\multicolumn{1}{l}{89.9/\textbf{100.0}}
\\
&\multicolumn{1}{l}{Gaussian}
&\multicolumn{1}{l}{41.57/\textbf{0.0}}
&\multicolumn{1}{l}{22.78/\textbf{0.0}}
&\multicolumn{1}{l}{66.29/\textbf{100.0}}
&\multicolumn{1}{l}{79.97/\textbf{100.0}}
&\multicolumn{1}{l}{53.53/\textbf{100.0}}
\\
&\multicolumn{1}{l}{Uniform}
&\multicolumn{1}{l}{42.2/\textbf{0.0}}
&\multicolumn{1}{l}{23.56/\textbf{0.0}}
&\multicolumn{1}{l}{72.13/\textbf{100.0}}
&\multicolumn{1}{l}{82.32/\textbf{100.0}}
&\multicolumn{1}{l}{58.33/\textbf{100.0}}
\\
  \bottomrule
  \end{tabular}
\end{table*}
\begin{table*}
  \caption{Comparison between our approach (without preprocessing) and ODIN (without preprocessing) for the CIFAR-10 and CIFAR-100 datasets. $\downarrow$ indicates that lower values are better, whereas $\uparrow$ indicates that higher values are better. \textbf{Bold} indicates the best score.}
  \label{cifar_odin}
  \centering
  \begin{tabular}{lllllll}
    \toprule
    \shortstack{ID\\model}     & OOD     & \shortstack{FPR at\\ 95\% TPR} $ \downarrow$ & \shortstack{Detection \\ error$\downarrow$}  & AUROC$\uparrow$  & \shortstack {AUPR\\Out$\uparrow$}  & \shortstack{AUPR\\In$\uparrow$} \\
    \cmidrule(l){3-7}
    \multicolumn{5}{r}{ODIN / ours (both without preprocessing)}                   \\
    \midrule
\multirow{4}{*}{\shortstack{CIFAR10\\VGG16}}
&\multicolumn{1}{l}{TinyImagenet}
&\multicolumn{1}{l}{33.18/\textbf{25.6}}
&\multicolumn{1}{l}{19.08/\textbf{15.29}}
&\multicolumn{1}{l}{90.91/\textbf{95.91}}
&\multicolumn{1}{l}{91.58/\textbf{95.21}}
&\multicolumn{1}{l}{89.6/\textbf{96.52}}
\\
&\multicolumn{1}{l}{LSUN}
&\multicolumn{1}{l}{20.75/\textbf{8.75}}
&\multicolumn{1}{l}{12.86/\textbf{6.87}}
&\multicolumn{1}{l}{94.42/\textbf{98.21}}
&\multicolumn{1}{l}{95.38/\textbf{98.1}}
&\multicolumn{1}{l}{93.17/\textbf{98.36}}
\\
&\multicolumn{1}{l}{iSUN}
&\multicolumn{1}{l}{22.5/\textbf{10.66}}
&\multicolumn{1}{l}{13.74/\textbf{7.82}}
&\multicolumn{1}{l}{93.77/\textbf{98.04}}
&\multicolumn{1}{l}{94.82/\textbf{97.93}}
&\multicolumn{1}{l}{92.45/\textbf{98.21}}
\\
&\multicolumn{1}{l}{SVHN}
&\multicolumn{1}{l}{23.95/\textbf{8.46}}
&\multicolumn{1}{l}{14.46/\textbf{6.7}}
&\multicolumn{1}{l}{92.07/\textbf{97.32}}
&\multicolumn{1}{l}{93.82/\textbf{97.95}}
&\multicolumn{1}{l}{88.7/\textbf{95.64}}
\\
&\multicolumn{1}{l}{Gaussian}
&\multicolumn{1}{l}{26.95/\textbf{0.0}}
&\multicolumn{1}{l}{15.9/\textbf{0.0}}
&\multicolumn{1}{l}{85.36/\textbf{100.0}}
&\multicolumn{1}{l}{90.61/\textbf{100.0}}
&\multicolumn{1}{l}{73.12/\textbf{100.0}}
\\
&\multicolumn{1}{l}{Uniform}
&\multicolumn{1}{l}{50.5/\textbf{0.0}}
&\multicolumn{1}{l}{27.74/\textbf{0.0}}
&\multicolumn{1}{l}{70.1/\textbf{100.0}}
&\multicolumn{1}{l}{79.7/\textbf{100.0}}
&\multicolumn{1}{l}{57.25/\textbf{100.0}}
\\
\midrule
\multirow{4}{*}{\shortstack{CIFAR10\\ResNet-V1-44}}
&\multicolumn{1}{l}{TinyImagenet}
&\multicolumn{1}{l}{29.75/\textbf{7.93}}
&\multicolumn{1}{l}{17.37/\textbf{6.45}}
&\multicolumn{1}{l}{91.67/\textbf{98.32}}
&\multicolumn{1}{l}{92.89/\textbf{97.83}}
&\multicolumn{1}{l}{89.61/\textbf{98.65}}
\\
&\multicolumn{1}{l}{LSUN}
&\multicolumn{1}{l}{17.43/\textbf{2.49}}
&\multicolumn{1}{l}{11.2/\textbf{3.73}}
&\multicolumn{1}{l}{95.19/\textbf{99.14}}
&\multicolumn{1}{l}{96.14/\textbf{98.99}}
&\multicolumn{1}{l}{93.6/\textbf{99.29}}
\\
&\multicolumn{1}{l}{iSUN}
&\multicolumn{1}{l}{19.0/\textbf{7.38}}
&\multicolumn{1}{l}{12.0/\textbf{6.19}}
&\multicolumn{1}{l}{94.67/\textbf{98.57}}
&\multicolumn{1}{l}{95.7/\textbf{98.31}}
&\multicolumn{1}{l}{93.08/\textbf{98.81}}
\\
&\multicolumn{1}{l}{SVHN}
&\multicolumn{1}{l}{15.12/\textbf{4.44}}
&\multicolumn{1}{l}{10.06/\textbf{4.66}}
&\multicolumn{1}{l}{96.25/\textbf{98.68}}
&\multicolumn{1}{l}{96.8/\textbf{98.78}}
&\multicolumn{1}{l}{95.43/\textbf{98.12}}
\\
&\multicolumn{1}{l}{Gaussian}
&\multicolumn{1}{l}{85.7/\textbf{0.0}}
&\multicolumn{1}{l}{45.32/\textbf{0.0}}
&\multicolumn{1}{l}{34.42/\textbf{100.0}}
&\multicolumn{1}{l}{50.86/\textbf{100.0}}
&\multicolumn{1}{l}{38.68/\textbf{100.0}}
\\
&\multicolumn{1}{l}{Uniform}
&\multicolumn{1}{l}{38.61/\textbf{0.0}}
&\multicolumn{1}{l}{21.78/\textbf{0.0}}
&\multicolumn{1}{l}{77.72/\textbf{100.0}}
&\multicolumn{1}{l}{85.46/\textbf{100.0}}
&\multicolumn{1}{l}{63.84/\textbf{100.0}}
\\
\midrule
\multirow{4}{*}{\shortstack{CIFAR100\\VGG16}}
&\multicolumn{1}{l}{TinyImagenet}
&\multicolumn{1}{l}{53.49/\textbf{51.87}}
&\multicolumn{1}{l}{29.24/\textbf{28.43}}
&\multicolumn{1}{l}{81.29/\textbf{91.14}}
&\multicolumn{1}{l}{83.73/\textbf{89.0}}
&\multicolumn{1}{l}{76.85/\textbf{92.73}}
\\
&\multicolumn{1}{l}{LSUN}
&\multicolumn{1}{l}{49.32/\textbf{28.14}}
&\multicolumn{1}{l}{27.16/\textbf{16.55}}
&\multicolumn{1}{l}{81.47/\textbf{95.36}}
&\multicolumn{1}{l}{84.74/\textbf{94.62}}
&\multicolumn{1}{l}{75.92/\textbf{96.04}}
\\
&\multicolumn{1}{l}{iSUN}
&\multicolumn{1}{l}{53.33/\textbf{31.27}}
&\multicolumn{1}{l}{29.17/\textbf{18.13}}
&\multicolumn{1}{l}{80.55/\textbf{94.66}}
&\multicolumn{1}{l}{83.75/\textbf{93.98}}
&\multicolumn{1}{l}{74.85/\textbf{95.42}}
\\
&\multicolumn{1}{l}{SVHN}
&\multicolumn{1}{l}{49.3/\textbf{22.23}}
&\multicolumn{1}{l}{27.13/\textbf{13.6}}
&\multicolumn{1}{l}{82.31/\textbf{92.5}}
&\multicolumn{1}{l}{85.3/\textbf{94.23}}
&\multicolumn{1}{l}{77.81/\textbf{87.44}}
\\
&\multicolumn{1}{l}{Gaussian}
&\multicolumn{1}{l}{81.97/\textbf{0.0}}
&\multicolumn{1}{l}{42.85/\textbf{0.0}}
&\multicolumn{1}{l}{27.22/\textbf{100.0}}
&\multicolumn{1}{l}{50.36/\textbf{100.0}}
&\multicolumn{1}{l}{36.68/\textbf{100.0}}
\\
&\multicolumn{1}{l}{Uniform}
&\multicolumn{1}{l}{82.37/\textbf{0.0}}
&\multicolumn{1}{l}{43.09/\textbf{0.0}}
&\multicolumn{1}{l}{26.78/\textbf{100.0}}
&\multicolumn{1}{l}{50.06/\textbf{100.0}}
&\multicolumn{1}{l}{36.57/\textbf{100.0}}
\\
\midrule
\multirow{4}{*}{\shortstack{CIFAR100\\ResNet-V1-44}}
&\multicolumn{1}{l}{TinyImagenet}
&\multicolumn{1}{l}{52.91/\textbf{20.45}}
&\multicolumn{1}{l}{28.94/\textbf{12.7}}
&\multicolumn{1}{l}{82.26/\textbf{96.77}}
&\multicolumn{1}{l}{84.43/\textbf{96.32}}
&\multicolumn{1}{l}{78.31/\textbf{97.24}}
\\
&\multicolumn{1}{l}{LSUN}
&\multicolumn{1}{l}{42.07/\textbf{19.68}}
&\multicolumn{1}{l}{23.53/\textbf{12.33}}
&\multicolumn{1}{l}{86.44/\textbf{96.81}}
&\multicolumn{1}{l}{88.6/\textbf{96.55}}
&\multicolumn{1}{l}{82.7/\textbf{97.21}}
\\
&\multicolumn{1}{l}{iSUN}
&\multicolumn{1}{l}{46.58/\textbf{21.92}}
&\multicolumn{1}{l}{25.78/\textbf{13.46}}
&\multicolumn{1}{l}{84.32/\textbf{96.37}}
&\multicolumn{1}{l}{86.87/\textbf{96.14}}
&\multicolumn{1}{l}{80.07/\textbf{96.81}}
\\
&\multicolumn{1}{l}{SVHN}
&\multicolumn{1}{l}{34.33/\textbf{12.15}}
&\multicolumn{1}{l}{19.64/\textbf{8.55}}
&\multicolumn{1}{l}{88.98/\textbf{96.95}}
&\multicolumn{1}{l}{91.14/\textbf{97.4}}
&\multicolumn{1}{l}{84.98/\textbf{95.2}}
\\
&\multicolumn{1}{l}{Gaussian}
&\multicolumn{1}{l}{98.71/\textbf{0.0}}
&\multicolumn{1}{l}{49.69/\textbf{0.0}}
&\multicolumn{1}{l}{2.83/\textbf{100.0}}
&\multicolumn{1}{l}{31.77/\textbf{100.0}}
&\multicolumn{1}{l}{31.16/\textbf{100.0}}
\\
&\multicolumn{1}{l}{Uniform}
&\multicolumn{1}{l}{88.34/\textbf{0.0}}
&\multicolumn{1}{l}{45.86/\textbf{0.0}}
&\multicolumn{1}{l}{19.23/\textbf{100.0}}
&\multicolumn{1}{l}{43.7/\textbf{100.0}}
&\multicolumn{1}{l}{34.6/\textbf{100.0}}
\\

  \bottomrule
  \end{tabular}
\end{table*}

\begin{table*}
  \caption{Comparison between our approach (without preprocessing) and MD (without preprocessing) for the CIFAR-10 and CIFAR-100 datasets. $\downarrow$ indicates that lower values are better, whereas $\uparrow$ indicates that higher values are better. \textbf{Bold} indicates the best score.}
  \label{cifar_md}
  \centering
  \begin{tabular}{lllllll}
    \toprule
    \shortstack{ID\\model}     & OOD     & \shortstack{FPR at\\ 95\% TPR} $ \downarrow$ & \shortstack{Detection \\ error$\downarrow$}  & AUROC$\uparrow$  & \shortstack {AUPR\\Out$\uparrow$}  & \shortstack{AUPR\\In$\uparrow$} \\
    \cmidrule(l){3-7}
    \multicolumn{5}{r}{MD / ours (both without preprocessing)}                   \\
    \midrule
\multirow{4}{*}{\shortstack{CIFAR10\\VGG16}}
&\multicolumn{1}{l}{TinyImagenet}
&\multicolumn{1}{l}{31.94/\textbf{25.6}}
&\multicolumn{1}{l}{18.47/\textbf{15.29}}
&\multicolumn{1}{l}{90.46/\textbf{95.91}}
&\multicolumn{1}{l}{91.42/\textbf{95.21}}
&\multicolumn{1}{l}{88.68/\textbf{96.52}}
\\
&\multicolumn{1}{l}{LSUN}
&\multicolumn{1}{l}{21.46/\textbf{8.75}}
&\multicolumn{1}{l}{13.21/\textbf{6.87}}
&\multicolumn{1}{l}{93.66/\textbf{98.21}}
&\multicolumn{1}{l}{94.77/\textbf{98.1}}
&\multicolumn{1}{l}{92.27/\textbf{98.36}}
\\
&\multicolumn{1}{l}{iSUN}
&\multicolumn{1}{l}{23.71/\textbf{10.66}}
&\multicolumn{1}{l}{14.34/\textbf{7.82}}
&\multicolumn{1}{l}{92.87/\textbf{98.04}}
&\multicolumn{1}{l}{94.17/\textbf{97.93}}
&\multicolumn{1}{l}{91.32/\textbf{98.21}}
\\
&\multicolumn{1}{l}{SVHN}
&\multicolumn{1}{l}{33.48/\textbf{8.46}}
&\multicolumn{1}{l}{19.22/\textbf{6.7}}
&\multicolumn{1}{l}{87.57/\textbf{97.32}}
&\multicolumn{1}{l}{90.4/\textbf{97.95}}
&\multicolumn{1}{l}{79.89/\textbf{95.64}}
\\
&\multicolumn{1}{l}{Gaussian}
&\multicolumn{1}{l}{13.65/\textbf{0.0}}
&\multicolumn{1}{l}{8.89/\textbf{0.0}}
&\multicolumn{1}{l}{93.49/\textbf{100.0}}
&\multicolumn{1}{l}{95.75/\textbf{100.0}}
&\multicolumn{1}{l}{89.29/\textbf{100.0}}
\\
&\multicolumn{1}{l}{Uniform}
&\multicolumn{1}{l}{19.05/\textbf{0.0}}
&\multicolumn{1}{l}{11.51/\textbf{0.0}}
&\multicolumn{1}{l}{87.98/\textbf{100.0}}
&\multicolumn{1}{l}{92.79/\textbf{100.0}}
&\multicolumn{1}{l}{75.67/\textbf{100.0}}
\\
\midrule
\multirow{4}{*}{\shortstack{CIFAR10\\ResNet-V1-44}}
&\multicolumn{1}{l}{TinyImagenet}
&\multicolumn{1}{l}{81.25/\textbf{7.93}}
&\multicolumn{1}{l}{43.08/\textbf{6.45}}
&\multicolumn{1}{l}{61.35/\textbf{98.32}}
&\multicolumn{1}{l}{64.74/\textbf{97.83}}
&\multicolumn{1}{l}{56.26/\textbf{98.65}}
\\
&\multicolumn{1}{l}{LSUN}
&\multicolumn{1}{l}{80.59/\textbf{2.49}}
&\multicolumn{1}{l}{42.78/\textbf{3.73}}
&\multicolumn{1}{l}{56.82/\textbf{99.14}}
&\multicolumn{1}{l}{62.99/\textbf{98.99}}
&\multicolumn{1}{l}{50.68/\textbf{99.29}}
\\
&\multicolumn{1}{l}{iSUN}
&\multicolumn{1}{l}{82.21/\textbf{7.38}}
&\multicolumn{1}{l}{43.6/\textbf{6.19}}
&\multicolumn{1}{l}{56.69/\textbf{98.57}}
&\multicolumn{1}{l}{62.04/\textbf{98.31}}
&\multicolumn{1}{l}{51.19/\textbf{98.81}}
\\
&\multicolumn{1}{l}{SVHN}
&\multicolumn{1}{l}{65.94/\textbf{4.44}}
&\multicolumn{1}{l}{35.47/\textbf{4.66}}
&\multicolumn{1}{l}{76.25/\textbf{98.68}}
&\multicolumn{1}{l}{78.12/\textbf{98.78}}
&\multicolumn{1}{l}{71.09/\textbf{98.12}}
\\
&\multicolumn{1}{l}{Gaussian}
&\multicolumn{1}{l}{0.0/\textbf{0.0}}
&\multicolumn{1}{l}{0.0/\textbf{0.0}}
&\multicolumn{1}{l}{100.0/\textbf{100.0}}
&\multicolumn{1}{l}{100.0/\textbf{100.0}}
&\multicolumn{1}{l}{100.0/\textbf{100.0}}
\\
&\multicolumn{1}{l}{Uniform}
&\multicolumn{1}{l}{0.0/\textbf{0.0}}
&\multicolumn{1}{l}{0.01/\textbf{0.0}}
&\multicolumn{1}{l}{100.0/\textbf{100.0}}
&\multicolumn{1}{l}{100.0/\textbf{100.0}}
&\multicolumn{1}{l}{100.0/\textbf{100.0}}
\\
\midrule
\multirow{4}{*}{\shortstack{CIFAR100\\VGG16}}
&\multicolumn{1}{l}{TinyImagenet}
&\multicolumn{1}{l}{58.21/\textbf{51.87}}
&\multicolumn{1}{l}{31.6/\textbf{28.43}}
&\multicolumn{1}{l}{79.09/\textbf{91.14}}
&\multicolumn{1}{l}{81.31/\textbf{89.0}}
&\multicolumn{1}{l}{74.68/\textbf{92.73}}
\\
&\multicolumn{1}{l}{LSUN}
&\multicolumn{1}{l}{55.22/\textbf{28.14}}
&\multicolumn{1}{l}{30.06/\textbf{16.55}}
&\multicolumn{1}{l}{80.08/\textbf{95.36}}
&\multicolumn{1}{l}{82.59/\textbf{94.62}}
&\multicolumn{1}{l}{75.71/\textbf{96.04}}
\\
&\multicolumn{1}{l}{iSUN}
&\multicolumn{1}{l}{58.72/\textbf{31.27}}
&\multicolumn{1}{l}{31.86/\textbf{18.13}}
&\multicolumn{1}{l}{78.86/\textbf{94.66}}
&\multicolumn{1}{l}{81.16/\textbf{93.98}}
&\multicolumn{1}{l}{74.17/\textbf{95.42}}
\\
&\multicolumn{1}{l}{SVHN}
&\multicolumn{1}{l}{53.26/\textbf{22.23}}
&\multicolumn{1}{l}{29.11/\textbf{13.6}}
&\multicolumn{1}{l}{77.03/\textbf{92.5}}
&\multicolumn{1}{l}{81.99/\textbf{94.23}}
&\multicolumn{1}{l}{69.74/\textbf{87.44}}
\\
&\multicolumn{1}{l}{Gaussian}
&\multicolumn{1}{l}{51.73/\textbf{0.0}}
&\multicolumn{1}{l}{27.84/\textbf{0.0}}
&\multicolumn{1}{l}{58.68/\textbf{100.0}}
&\multicolumn{1}{l}{74.3/\textbf{100.0}}
&\multicolumn{1}{l}{48.91/\textbf{100.0}}
\\
&\multicolumn{1}{l}{Uniform}
&\multicolumn{1}{l}{67.74/\textbf{0.0}}
&\multicolumn{1}{l}{36.14/\textbf{0.0}}
&\multicolumn{1}{l}{42.55/\textbf{100.0}}
&\multicolumn{1}{l}{62.8/\textbf{100.0}}
&\multicolumn{1}{l}{41.71/\textbf{100.0}}
\\
\midrule
\multirow{4}{*}{\shortstack{CIFAR100\\ResNet-V1-44}}
&\multicolumn{1}{l}{TinyImagenet}
&\multicolumn{1}{l}{86.99/\textbf{20.45}}
&\multicolumn{1}{l}{45.99/\textbf{12.7}}
&\multicolumn{1}{l}{56.66/\textbf{96.77}}
&\multicolumn{1}{l}{59.31/\textbf{96.32}}
&\multicolumn{1}{l}{52.74/\textbf{97.24}}
\\
&\multicolumn{1}{l}{LSUN}
&\multicolumn{1}{l}{87.16/\textbf{19.68}}
&\multicolumn{1}{l}{46.06/\textbf{12.33}}
&\multicolumn{1}{l}{54.02/\textbf{96.81}}
&\multicolumn{1}{l}{58.24/\textbf{96.55}}
&\multicolumn{1}{l}{50.01/\textbf{97.21}}
\\
&\multicolumn{1}{l}{iSUN}
&\multicolumn{1}{l}{90.77/\textbf{21.92}}
&\multicolumn{1}{l}{47.88/\textbf{13.46}}
&\multicolumn{1}{l}{50.76/\textbf{96.37}}
&\multicolumn{1}{l}{53.47/\textbf{96.14}}
&\multicolumn{1}{l}{48.17/\textbf{96.81}}
\\
&\multicolumn{1}{l}{SVHN}
&\multicolumn{1}{l}{81.16/\textbf{12.15}}
&\multicolumn{1}{l}{43.07/\textbf{8.55}}
&\multicolumn{1}{l}{62.49/\textbf{96.95}}
&\multicolumn{1}{l}{65.85/\textbf{97.4}}
&\multicolumn{1}{l}{57.26/\textbf{95.2}}
\\
&\multicolumn{1}{l}{Gaussian}
&\multicolumn{1}{l}{0.0/\textbf{0.0}}
&\multicolumn{1}{l}{0.0/\textbf{0.0}}
&\multicolumn{1}{l}{100.0/\textbf{100.0}}
&\multicolumn{1}{l}{100.0/\textbf{100.0}}
&\multicolumn{1}{l}{100.0/\textbf{100.0}}
\\
&\multicolumn{1}{l}{Uniform}
&\multicolumn{1}{l}{0.04/\textbf{0.0}}
&\multicolumn{1}{l}{0.14/\textbf{0.0}}
&\multicolumn{1}{l}{99.99/\textbf{100.0}}
&\multicolumn{1}{l}{99.99/\textbf{100.0}}
&\multicolumn{1}{l}{99.97/\textbf{100.0}}
\\

  \bottomrule
  \end{tabular}
\end{table*}


\begin{table*}
  \caption{Hyper parameters used for different models in the experiments}
  \label{model_hyper_parameters}
  \centering
  \begin{tabular}{lcccccc}
    \toprule
    Model & Optimizer & Epochs  & \shortstack{Batch\\size} &  \shortstack{learning\\rate} &  \shortstack{Dropout\\layer} & \shortstack{Data\\ augmentation}\\
    \midrule
ResNet & Adam & 200 & 32 & 0.001 & No & Yes
\\
VGG-16 & SGD  & 250 & 128 & 0.1 & Yes & Yes 
\\
CNN (MNIST) & Adam & 100 & 5000 &  0.001 & Yes & Yes
\\
  \bottomrule
  \end{tabular}
\end{table*}

\end{document}